\documentclass{article}

\usepackage[final,nonatbib]{nips_2018}

\usepackage{microtype}
\usepackage{graphicx}
\usepackage{caption}
\usepackage{subcaption}
\usepackage{subfloat}
\usepackage{booktabs} 
\usepackage{xspace}
\usepackage{courier}
\usepackage{amsmath}
\usepackage{amsfonts}
\usepackage{array}
\usepackage{comment}
\usepackage{multirow}
\usepackage{url}
\usepackage{wrapfig}

\usepackage{enumitem}
\usepackage{lipsum}
\setitemize{noitemsep,topsep=2pt,parsep=0pt,partopsep=0pt,leftmargin=*}

\makeatletter
\renewcommand{\paragraph}{%
  \@startsection{paragraph}{4}%
  {\z@}{1ex \@plus 1ex \@minus .2ex}{-1em}%
  {\normalfont\normalsize\bfseries}%
}
\makeatother

\newcommand{\figlabel}[1]{\label{fig:#1}}
\newcommand{\figref}[1]{Figure~\ref{fig:#1}}

\newcommand{\system}{\textsc{Houdini}\xspace}

\newcommand{\secref}[1]{Section~\ref{sec:#1}}
\newcommand{\seclabel}[1]{\label{sec:#1}}

\usepackage[dvipsnames]{xcolor}
\newcommand{\maybecomment}[1]{}

\newcommand{\lazar}[1]{{\color{green}\maybecomment{From Lazar: #1}}}
\newcommand{\swarat}[1]{{\color{blue}\maybecomment{From Swarat: #1}}}

\newcommand{\error}[1]{#1\% CE}

\newcommand{\rmse}[1]{#1 RMSE}
\newcommand{\score}{\textbf{Error}\hspace{2em}\ }
\newcommand{\progcolwidth}{30em}

\title{\system: Lifelong Learning as Program Synthesis}

\author
{Lazar Valkov \\ University of Edinburgh \\ L.Valkov@sms.ed.ac.uk \And 
Dipak Chaudhari \\ Rice University \\ dipakc@rice.edu \And
Akash Srivastava \\ University of Edinburgh \\ Akash.Srivastava@ed.ac.uk \And 
Charles Sutton \\ University of Edinburgh \\ csutton@inf.ed.ac.uk \And 
Swarat Chaudhuri \\ Rice University \\ swarat@rice.edu
}

\newcommand{\citet}[1]{\cite{#1}}
\newcommand{\citep}[1]{\cite{#1}}

\begin{document}

\maketitle
    \begin{abstract}
        We present a {\em neurosymbolic framework} for the lifelong learning
        of algorithmic tasks that mix perception and procedural
        reasoning. Reusing high-level concepts across domains and learning
        complex procedures are key challenges in lifelong learning. We
        show that a {\em program synthesis} approach that combines gradient descent with combinatorial search over programs can be a more effective response to these
        challenges than purely neural methods. Our framework, called \system, represents neural networks as strongly typed,
        differentiable functional programs that use 
        symbolic higher-order combinators to compose a library of neural
        functions. 
        Our learning algorithm consists of: (1) a symbolic program
        synthesizer that performs a type-directed search over
        parameterized programs, and decides on the 
        library functions to
        reuse, and the architectures to combine them, while
        learning a sequence of tasks; and (2) a neural module that
        trains these programs using stochastic gradient descent.
       We evaluate \system on  three benchmarks that combine perception with the algorithmic tasks of counting, summing, and shortest-path computation. 
        Our experiments show that \system transfers high-level
        concepts more effectively than traditional transfer learning
        and progressive neural networks, and that the typed representation of networks  significantly accelerates the search. 
    \end{abstract}

    \section{Introduction}

{\em Differentiable programming
languages}~\cite{autograd, paszke2017automatic,BosnjakRNR17,GauntBSKKTT16,BunelDMKT16, chainer_learningsys2015,eager}
have recently emerged as a powerful approach to the task of engineering deep
learning systems.
These languages are syntactically similar to, and often direct extensions of, 
traditional programming
languages. However, programs in these languages are differentiable in their
parameters, permitting gradient-based parameter learning. The
framework of differentiable languages has proven especially powerful for representing
architectures that have input-dependent structure, such as deep
networks over trees \cite{socher2013recursive,allamanis17symbolic} and
graphs \cite{li16gated,kipf17graph}. 

A recent paper by Gaunt et al.
~\citet{GauntBKT17} points to another key appeal of high-level
differentiable languages: they
facilitate {\em transfer} across learning tasks.  The paper
gives a language called {\sc Neural Terpret} (NTPT) in which
programs can invoke a library of trainable neural networks as
subroutines. The parameters of these ``library functions'' are learned
along with the code that calls them. The modularity of the language allows
knowledge transfer, as a library function trained on a task
can be reused in later tasks. In contrast to standard methods 
for transfer learning in deep networks, which re-use the first
few layers of the network,
neural libraries have the potential to enable reuse
of higher, more abstract levels of the network,
in what could be called \emph{high-level transfer}.
In spite of its promise, though, inferring the structure of
differentiable programs is a fundamentally hard problem.
While NTPT
and its predecessor {\sc Terpret}~\cite{GauntBSKKTT16} 
allow some aspects of the program structure to be induced, a detailed
hand-written template of the program is required for even the simplest
tasks.

In this paper, we show that algorithmic ideas from {\em program synthesis} can help overcome this difficulty. The goal in program
synthesis~\citep{sygus,Solar-Lezama13,FeserCD15} is to discover 
programs (represented as terms following a specified syntax) that accomplish a given task. 
Many symbolic algorithms for the problem have been proposed in the recent past~\cite{gulwani-survey}.  
These algorithms can often outperform
purely neural approaches on procedural tasks~\citep{GauntBSKKTT16}. 
A key insight behind our approach is that these methods naturally complement 
stochastic gradient descent (SGD)
in the induction of differentiable programs: while SGD is 
an effective way of learning program parameters, each step in a symbolic
search can explore large changes to the program structure.

A second feature of our work is a representation of programs in a {\em typed functional language}. Such a representation is natural because
functional combinators can compactly describe many common neural architectures
\cite{olahblog}. For
example, {\tt fold} combinators can describe recurrent
neural networks,  
and convolution over data structures such as lists and graphs can also
be naturally expressed as functional combinators. Such representations also facilitate search, as they tend to be more canonical, and as the type system can help prune infeasible programs early on in the search~\cite{FeserCD15,myth}. 

Concretely, we present a {\em neurosymbolic} learning framework,  
called \system, which is to our knowledge 
the first method to use symbolic methods to induce  
differentiable programs. In \system, a program synthesizer is used to search over networks
described as strongly typed functional programs,
whose parameters are then tuned end-to-end using gradient descent. 
Programs in \system specify the architecture of the network, 
by using functional combinators to express the network's connections,
and can also facilitate learning transfer, by letting the synthesizer
choose among library functions. \system is flexible about
how the program synthesizer is implemented:  here, we
use and compare two implementations, one performing top-down,
type-directed enumeration with a preference for simpler programs, and the other based on a type-directed evolutionary
 algorithm. The implementation for \system is available online~\cite{houdinirepo}.

We evaluate \system in the setting of 
{\em lifelong learning}~\citep{ThrunM95}, in which a model is trained on a
series of tasks, and each training round is expected to benefit from
previous rounds of learning.  
Two challenges in lifelong learning are \emph{catastrophic forgetting}, in which later tasks overwrite what
has been learned from earlier tasks, and \emph{negative transfer}, in
  which attempting to use information from earlier tasks hurts
performance on later tasks. Our use of a neural library avoids both problems, as the library allows us to
freeze and selectively re-use portions of networks that have been
successful. 

Our experimental benchmarks combine perception with algorithmic tasks such as counting, summing, and shortest-path computation. Our programming language allows parsimonious representation for such tasks, as the 
combinators used to describe network structure can also be used to compactly express rich algorithmic operations. 
Our experiments show that
\system can learn nontrivial programs for these tasks. 
For example, on a task of computing least-cost paths in a grid of 
images, \system discovers an algorithm that has the structure
of the Bellman-Ford shortest path
algorithm~\citep{bellman1958routing}, but uses a learned neural
function that approximates the algorithm's ``relaxation''
step. Second, our results indicate that the modular representation used in \system allows it to transfer high-level concepts and avoid negative transfer. We demonstrate that \system offers greater transfer than progressive
neural networks~\cite{rusu2016progressive} and 
traditional ``low-level'' transfer~\cite{yosinski14}, in which early network layers are inherited from previous tasks.
Third, we show that the
use of a higher-level, typed language is critical to scaling the
search for programs.

The contributions of this paper are threefold. First, we propose the use of symbolic program synthesis in transfer and lifelong learning. Second, we introduce a specific representation of neural networks as typed functional programs, whose types contain rich information such as tensor dimensions, and show how to leverage this representation in program synthesis. Third, we show that the modularity inherent in typed functional programming allows the method to transfer high-level modules, to avoid negative transfer and is capable of few-shot learning.

\paragraph{Related Work.} 



\system builds on a known insight from
program synthesis~\citet{gulwani-survey}: that functional
representations and type-based pruning can be used to accelerate search over programs~\citep{FeserCD15,myth,LeG14}. However, most prior efforts on
program synthesis are purely symbolic and driven by the Boolean goals. \system repurposes these methods for
an optimization setting, coupling them with gradient-based learning.
A few recent approaches to program synthesis do combine symbolic and
neural methods~\citep{robustfill,balog2016deepcoder,
ellis2017graphics,parisotto16,kalyan18}.  
Unlike our work, these methods do not synthesize differentiable
programs. The only exception is  NTPT~\citet{GauntBKT17},
which neither considers a functional language nor a neurosymbolic
search. Another recent method that creates a neural library is 
progress-and-compress \cite{progress-compress},  but this method is restricted to feedforward networks and low-level transfer. It is based on progressive networks \cite{rusu2016progressive}, a method for lifelong learning based on low-level transfer, in which lateral connections are added between the networks for the all the previous tasks and the new task.

 \emph{Neural module networks} (NMNs)
\citep{andreas2016neural, hu2017learning}
 select and arrange
 reusable neural modules into a program-like network, for
visual question answering.  The major
difference from our work is that NMNs require a natural
language input to guide decisions about which modules to combine.  \system works
without this additional supervision.         
Also, \system can be seen to 
perform a form of
\emph{neural architecture search}.  Such search has been studied extensively, using approaches such as reinforcement learning,
evolutionary computation, and best-first search
\citep{zoph17,liu17,real17a,zhong17}. 
\system operates at a higher level of abstraction, combining
entire networks that have been trained previously, rather than optimizing
over lower-level decisions such as the 
width of convolutional filters, the details
of the gating mechanism, and so on.
\system is distinct in
its use of functional programming to represent
architectures compactly and abstractly, and in its extensive use of
types in accelerating search.

    \section{The \system Programming Language}\seclabel{lang}

    \newcommand{\algo}{\textsc{Learn}\xspace}
\newcommand{\symmod}{\textsc{Generate}\xspace}
\newcommand{\neuromod}{\textsc{Tune}\xspace}
\newcommand{\D}{\mathcal{D}}
\newcommand{\Lib}{\ensuremath{\mathcal{L}}}
\newcommand{\List}{\mathtt{list}}

\newcommand{\tensor}[2]{Tensor[#1]\langle#2\rangle}
\newcommand{\graph}[2]{Graph[#1]\langle#2\rangle}
\newcommand{\tree}[1]{Tree[#1]}
\newcommand{\lst}[1]{List[#1]}
\newcommand{\mto}{\rightarrow}
\newcommand{\gbar}{\ \,| \ \,}
\newcommand{\hole}[1]{\langle\hspace{-0.2em}\langle #1 \rangle\hspace{-0.2em}\rangle}

\newcommand{\slangle}[1]{\langle#1\rangle}
\newcommand{\fop}{\oplus}
\newcommand{\hop}{\otimes}
\newcommand{\lambdat}[2]{\lambda\ #1.\ #2}
\newcommand{\sayit}[1]{}
\newcommand{\hilight}[1]{{\color{red} #1}}

\newcommand{\id}[1]{\mathit{#1}}

\newcommand{\Tensor}{\mathtt{Tensor}}
\newcommand{\ADT}{\mathit{ADT}}

\newcommand{\mapc}{\mathbf{map}}
\newcommand{\foldc}{\mathbf{fold}}
\newcommand{\convc}{\mathbf{conv}}

\newcommand{\composec}{\mathbf{compose}}
\newcommand{\repeatc}{\mathbf{repeat}}
\newcommand{\zerosc}{\mathbf{zeros}}

\newcommand{\mapg}{\mathbf{map\_g}}
\newcommand{\foldg}{\mathbf{fold\_g}}
\newcommand{\convg}{\mathbf{conv\_g}}

\newcommand{\mapl}{\mathbf{map\_l}}
\newcommand{\foldl}{\mathbf{fold\_l}}
\newcommand{\convl}{\mathbf{conv\_l}}

\newcommand{\zug}[1]{\langle #1 \rangle}

\system consists of two components. The first is a
typed, higher-order, functional language of differentiable programs. The
second is a learning procedure split into a symbolic module and a neural module. Given a task
specified by a set of training examples,
the symbolic module enumerates parameterized programs in the \system language. The neural module uses
gradient descent to find optimal parameters for
synthesized programs; it also assesses the quality of solutions and decides
whether an adequately performant solution has been
discovered. 

The design of the language is based on three ideas: 
\begin{itemize} 
\item The ubiquitous use of {\em function composition} to glue
  together different networks. 

\item The use of {\em higher-order combinators} such as $\mapc$ and
  $\foldc$ to uniformly represent neural architectures as well as
  patterns of recursion in procedural tasks. 

\item The use of a strong {\em type discipline} to distinguish between neural
  computations over different forms of data, and to avoid generating 
  provably incorrect programs during symbolic exploration.
\end{itemize}

\figref{language} shows the grammar for the \system
language. 
Here, $\tau$ denotes types and $e$ denotes programs. 
Now we elaborate
on the various language constructs. 

\begin{figure}
\small 
{\bf Types} $\mathbf{\tau}$:\smallskip \\
$
\begin{array}{llllll}
\tau & ::= & \id{Atom} \mid \id{ADT} \mid F \qquad &
\id{Atom} &::=& \mathtt{bool} \mid \mathtt{real} \smallskip\\
\id{TT} &::= &  \Tensor\zug{\id{Atom}}[m_1][m_2]\dots[m_k] \qquad &
\id{ADT} & ::= & \id{TT} \mid \alpha\zug{\id{TT}} \smallskip \\
F & ::= & \id{ADT} \mid F_1 \mto F_2 & .
\end{array}
$ \medskip \\
{\bf Programs } $\mathbf{e}$ {\bf over library $\Lib$}:\qquad
$
\begin{array}{lll}
e & ::= 
  \fop_w:\tau_0 \mid e_0 \circ e_1 \mid
          {\mapc_{\alpha}\ e } \mid {\foldc_{\alpha}~e} \mid 
    {\convc_{\alpha}\ e }.
\end{array}
$
\caption{Syntax of the \system language. Here,
  $\alpha$ is an ADT, e.g., $\mathtt{list}$; $m_1,\dots, m_k \ge 1$ define the shape of a
  tensor; $F_1 \mto F_2$ is a function type; 
  $\fop_w \in \Lib$ is a neural
  library function that has type $\tau_0$ and parameters $w$; and $\circ$ is the
  composition operator. $\mapc$, $\foldc$, and $\convc$ are
  defined below.}\figlabel{language}
\vspace{-0.1in}
\end{figure}


\paragraph{Types.} 
The ``atomic'' data types in \system are booleans ({\tt bool}) and
reals. For us, {\tt bool} is relaxed into a real value in $\lbrack 0, 1 \rbrack,$ which for example, allows the type system
to track if a vector has been passed
through a sigmoid.
{\em Tensors} over these types are also permitted. We have a
distinct type $\Tensor \zug{\id{Atom}}[m_1][m_2]\dots[m_k]$ for
tensors  of shape $m_1 \times \dots \times m_k$ whose elements have
atomic type $\id{Atom}$. (The dimensions $m_1,\dots, m_k$, as well as $k$ itself, are bounded to keep
the set of types finite.)
We also have function types $F_1 \rightarrow F_2$, and abstract data types (ADTs) $\alpha\zug{\id{TT}}$
parameterized by a tensor type $\id{TT}$. Our current
implementation supports two kinds of ADTs: $\mathtt{list}\zug{\id{TT}}$, lists with
elements of tensor type $\id{TT}$,
and $\mathtt{graph}\zug{\id{TT}}$,
 graphs whose nodes have values of tensor type $\id{TT}$. 

\paragraph{Programs.} 
The fundamental operation in \system is {\em function composition}.
A composition operation can involve   
functions $\fop_w$, parameterized by weights $w$ and
implemented by neural networks, drawn from a library $\Lib$. It can also involve a set of symbolic higher-order combinators that are guaranteed to preserve end-to-end differentiability and used to 
implement high-level network architectures. Specifically, we allow the following three families of combinators. The first two are standard constructs for functional languages,
whereas the third is introduced specifically for deep models.
\begin{itemize}
\item Map combinators $\mapc_{\alpha\zug{\tau}}$, for ADTs of the form 
  $\alpha\zug{\tau}$. Suppose $e$ is a function. The expression
$\mapc_{\mathtt{list}\zug{\tau}}~e$ is a function that, given a
list $[a_1, \dots, a_k]$, returns the list $[e(a_1), \dots, e(a_k)]$. 
The expression
$\mapc_{\mathtt{graph}_{\tau}}~e$ is a function that, given a
graph $G$ whose $i$-th node is labeled with a value $a_i$, returns a graph
that is identical to $G$, but whose $i$-th node is labeled by
$e(a_i)$.

\item Left-fold combinators $\foldc_{\alpha\zug{\tau}}$.  For a
  function $e$ and a term $z$, $\foldc_{\mathtt{list}\zug{{\tau}}}~e~z$ is the function
  that, given a list $[a_1, \dots, a_k]$, returns the value
  $(e~(e~\dots (e~(e~z~a_1)~a_2)\dots)~a_k)$. To define fold over a graph, we
  assume a linear order on the graph's nodes. Given $G$, the function
  $\foldc_{\mathtt{graph}\zug{{\tau}}}~e~z$ returns the fold over the
  list $[a_1,\dots, a_k]$, where $a_i$ is the value at the $i$-th node
  in this order.

\item Convolution combinators $\convc_{\alpha\zug{\tau}}$. 
Let $p > 0$ be a fixed constant. 
For a ``kernel" function $e$, 
$\convc_{\mathtt{list}\zug{{\tau}}}~e$ is the function that, given a list 
 $[a_1, \dots, a_k]$, returns the list
 $[a_1', \dots, a_k']$, where $a_i' = e~[a_{i-p}, \dots,
 a_i,\dots,a_{i+p}]$. (We define $a_j = a_1$ if
 $j < 1$, and $a_j = a_k$ if $j > k$.)
Given a graph $G$, 
the function
$\convc_{\mathtt{graph}\zug{{\tau}}}~e$  returns the graph $G'$ whose 
node $u$ contains the value 
$e~[a_{i_1},\dots,a_{i_m}]$, where
$a_{i_j}$ is the value stored in the $j$-th neighbor of $u$.

\end{itemize}

Every neural library function is assumed to be annotated with a
type. 
Using  programming language
techniques~\citep{pierce2002types}, 
\system assigns a type to 
each program $e$ whose subexpressions use types consistently (see supplementary material). 
If it is impossible to assign a type to $e$,
then $e$ is {\em type-inconsistent}. Note that complete \system programs do not have explicit variable names.
Thus, \system follows the
{\em point-free} style of functional
programming~\citep{Backus:1978:PLV:359576.359579}.
This style permits highly succinct representations of complex computations, which reduces the amount of enumeration needed during synthesis.

\begin{wrapfigure}{l}{1.6in}
\vspace{-0.75em}
\centering
\includegraphics[scale=0.3]{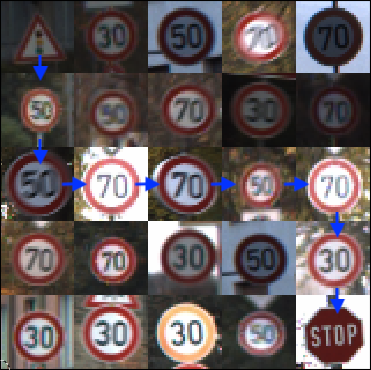}
\caption{A grid of 32x32x3 images from the GTSRB dataset
  ~\cite{Stallkamp2012}. The least-cost path from the top left to the bottom
  right node is marked.}\figlabel{streetsigns}
\vspace{-2em}
\end{wrapfigure}

\paragraph{\system for deep learning.}
The language has several properties
that are useful for specifying deep models. First, any complete \system
program $e$ 
is differentiable in the parameters $w$ of the neural library functions
used in $e$. Second, common deep architectures
can be compactly represented
in our language.
For example, 
deep feedforward networks can be represented by
$\fop_1 \circ \dots \circ \fop_k$, where each $\fop_i$ is a neural function,
and recurrent nets can be
expressed as $\foldc_{\mathtt{list}\zug{\tau}}~\fop~z$, where
$\fop$ is a neural function and $z$ is 
the initial state.  Graph convolutional
networks can be expressed as  
$\convc_{\mathtt{graph}\zug{\tau}}~\fop$. Going further, the
language can be easily extended to handle
bidirectional recurrent networks, attention
mechanisms, and so on.

\paragraph{Example: Shortest path in a grid of images.}
To show how \system can model tasks that mix
perception and procedural reasoning, we use an example that generalizes
the navigation task of Gaunt et al.~\citet{GauntBKT17}. 
Suppose we are given a grid of images (e.g., \figref{streetsigns}),
whose elements represent speed limits and are connected
horizontally and vertically, but not diagonally.
Passing through each node induces a penalty, which depends
on the node's speed limit, with lower 
speed limits having a higher penalty. 
The task is to predict the minimum cost $d(u)$ incurred while traveling from a fixed
starting point $\id{init}$ to every other node $u$. 

One way to compute these costs is using
the Bellman-Ford shortest-path algorithm~\citep{bellman1958routing}, whose 
$i$-th iteration computes an estimated minimum cost
$d_i(u)$ of travel to each node $u$ in the graph. The cost
estimates for the $(i + 1)$-th iteration are computed using a {\em
  relaxation} operation:
$d_{i+1}(u) := \min(d_i(u), \min_{v \in \id{Adj}(u)}d_i(v) + w(v))$, 
where $w(v)$ is the penalty and $\id{Adj}(u)$ the 
neighbors of $u$. 
As the update to $d_i(u)$ only depends on values at $u$
and its neighbors, the relaxation step can be represented as a
graph convolution.
As described in \secref{eval}, \system is able to discover an approximation of this program purely from data. The synthesized program uses a graph convolution, a graph map, a neural module that processes the images of speed limits, and a neural module that approximates the relaxation function.

    \section{Learning Algorithm}\seclabel{synth}


\begingroup 

Now we define our learning problem. For a \system
program $e_w$ parameterized by a vector $w$, let $e[w \mapsto v]$ be the
function for the specific parameter
vector $v$, i.e. by substituting $w$ by $v$ in $e$. Suppose we have a
library $\Lib$ of neural functions and a training set $\D$. As usual, we assume that $\D$ consists of i.i.d. samples from a distribution
$p_{\id{data}}$. We assume that $\D$ is properly typed, i.e., every
training instance  $(x_i, y_i) \in \D$ has the same type, which is known. This means
that we also know the type $\tau$ of our target function.
The goal in our learning problem
is to discover a program $e^*_w$ of type $\tau$, and values $v$ for $w$ such that
$
e^*_w[w \mapsto v] = \mathrm{argmin}_{e \in \mathit{Progs(\Lib)}, w \in
\mathbb{R}^n} (\mathbf{E}_{x \sim p_{{data}}}[l(e, \D, x)]),
$
where $\mathit{Progs(\Lib)}$ is the universe of all programs
over $\Lib$, and $l$ is a suitable loss function.

Our algorithm for this task consists of a
symbolic program synthesis module called \symmod and a gradient-based
optimization module called \neuromod. \symmod repeatedly generates
parameterized programs $e_w$ and ``proposes'' them to \neuromod. 
\neuromod uses stochastic gradient descent to find parameter values $v$ for $e_w$
that lead to the optimal value of the loss function on a training set,
and produces a program $e = e_w[w \mapsto v]$ with instantiated parameters.  The
final output of the algorithm is a program $e^*$, among all
programs $e$ as above, that leads to optimal loss on a validation set.

As each program proposed by \symmod is subjected to training,
\symmod can only afford to propose a small number of programs, out of the vast combinatorial space of all programs.  
Selecting these
programs is a difficult challenge.
We use and compare two strategies for this task. 
Now we 
sketch these strategies; for more details, see the supplementary material.

\begin{itemize}
\item The first strategy is {\em top-down iterative refinement},
  similar to the algorithm in the $\lambda^2$ program
  synthesizer~\citep{FeserCD15}. Here, the synthesis procedure
  iteratively generates a series of ``partial'' programs (i.e.,
  programs with missing code) over the library $\Lib$, starting with an ``empty'' program and
  ending with a complete program. A type inference procedure is used
  to rule out any partial program that is not type-safe. A cost
  heuristic is used to generate programs in an order of structural simplicity. Concretely, shorter programs are evaluated first.

\item The second method is an {\em evolutionary algorithm} inspired
  by work on functional genetic
  programming~\citep{briggs2006functional}. Here, we use selection,
  crossover, and mutation operators to evolve a population of
  programs over $\Lib$. Types play a key role: all programs in the population are ensured to be type-safe, and mutation and
  crossover only replace a subterm in a program with terms of the
  same type.
\end{itemize}

In both cases, the use of types vastly reduces the amount of
search that is needed, as the number of type-safe programs of a given size is a small fraction of the number of programs of that size. 
See
\secref{eval} for an experimental confirmation.

\paragraph{Lifelong Learning.}

A {\em lifelong
learning} setting is a sequence of related tasks $\D_1, \D_2, \ldots $,
where each task $\D_i$ has its own training and validation set.  Here, the learner is called repeatedly, once for each
task $\D_i$ using a neural library $\Lib_i$, returning a
best-performing program $e^*_i$ with parameters $v^*_i.$

We implement transfer learning simply by adding new modules to the
neural library after each call to the learner.  We add all neural
functions from $e^*_i$ back into the library, freezing their
parameters. More formally, let $\fop_{i1} \ldots \fop_{iK}$ be the
neural library functions which are called anywhere in $e^*_i$. Each
library function $\fop_{ik}$ has parameters $w_{ik}$, 
set to the value $v^*_{ik}$ by \neuromod.  The library for the
next task is then
$\Lib_{i+1} = \Lib_i \cup \{ \fop_{ik}\lbrack w_{ik} \mapsto v^*_{ik}
\rbrack \}.$ 
This process ensures that the parameter
vectors of $\fop_{ik}$ are frozen and can no longer be updated by
subsequent tasks. Thus, we prevent catastrophic forgetting by
design. Importantly, it is always possible for the synthesizer to
introduce ``fresh networks'' whose parameters have not been
pretrained. This is because the library always monotonically increases
over time, so that an original neural library function with untrained
parameters is still available.

This approach has the important implication that the set of neural library functions that the synthesizer uses is not fixed,
but continually evolving.
Because both trained and untrained
versions of the library functions
are available, this can be seen to permit {\em selective transfer},
meaning that depending on which version of the library function
\symmod chooses,
the learner has the option of using or not using previously learned
knowledge in a new task. This fact allows \system to avoid negative transfer.

\endgroup 

    \section{Evaluation}\seclabel{eval}


\newcommand{\isdigit}[1]{recognize\_digit($#1$)}
\newcommand{\recogdigit}[1]{recognize\_digit($#1$)}
\newcommand{\countdigit}[1]{count\_digit($#1$)}
\newcommand{\counttoy}[1]{count\_toy($#1$)}
\newcommand{\istoy}[1]{recognize\_toy($#1$)}
\newcommand{\classifydigit}{classify\_digit}
\newcommand{\sumdigits}{sum\_digits}
\newcommand{\regressspeed}{regress\_speed}
\newcommand{\regressmnist}{regress\_mnist}
\newcommand{\shortestpathmnist}{shortest\_path\_mnist}
\newcommand{\shortestpathstreet}{shortest\_path\_street}

Our evaluation studies four questions. First, we ask whether \system can learn nontrivial differentiable programs that combine perception and algorithmic reasoning. Second, we study 
if \system can  
transfer perceptual and algorithmic knowledge during lifelong
learning. We study three forms of transfer: \emph{low-level transfer}
of perceptual concepts across domains, \emph{high-level transfer} of
algorithmic concepts, and \emph{selective transfer} where the learning
method decides on which known concepts to reuse. Third, we study the value of our type-directed approach to synthesis. Fourth, we compare the performance of the top-down and evolutionary synthesis algorithms.

\lazar{Q: where do we first say that we Freeze the weights and therefore we don't have a problem with catastrophic forgetting.)}

\paragraph{Task Sequences.}
 
Each lifelong learning setting is a sequence of individual learning tasks. The full list of tasks is shown in Figure~\ref{fig:tasks}.  These tasks include object recognition tasks over three data sets: MNIST~\cite{lecun1998gradient},  NORB~\cite{lecun2004learning}, and the GTSRB data set of images of traffic signs~\cite{Stallkamp2012}.  
\swarat{All images are pre-processed to have zero mean and standard deviation of one. Additionally, NORB images were resized to 28x28 in order to match the MNIST dimensionality. Finally, the GTSRB images were all re-sized to 32x32x3.}
In addition, we have three algorithmic tasks: \emph{counting} the number of instances of images of a certain class in a list of images; \emph{summing} a list of images of digits; and the \emph{shortest path} computation described in \secref{lang}. 

We combine these tasks into seven sequences. Three of these (CS1, SS, GS1) involve low-level transfer,
in which earlier tasks are perceptual tasks like recognizing digits, while 
later tasks introduce higher-level algorithmic problems.
Three other task sequences (CS2, CS3, GS2) involve higher-level transfer,
 in which  earlier tasks introduce a high-level concept, while
 later tasks require a learner to re-use this concept on different perceptual inputs. 
 For example, in CS2, once \countdigit{d_1} is learned for counting digits of class $d_1$, the synthesizer can learn to reuse this counting network on a new digit class $d_2$, even if the learning system 
 has never seen $d_2$ before. 
The graph task sequence GS1 also demonstrates that the graph
convolution combinator in \system allows learning of complex graph
algorithms and GS2 tests if high-level transfer can be performed with
this more complex task. Finally, we include a task sequence
LS that is
designed to evaluate our method on a task sequence that is
both longer and that lacks a favourable curriculum. 
The sequence LS was initially randomly generated, and then slightly amended in order to evaluate all lifelong learning concepts discussed.


\begin{figure}[t!]
\begin{minipage}[t]{0.34\textwidth}
    \small\raggedright
    \textbf{Individual tasks}\\[3pt]
    {\recogdigit{d}:} Binary classification of whether image contains a digit~$d \in \{ 0 \dots 9 \}$ \\[3pt]
    {\classifydigit:} Classify a digit into digit categories $(0-9)$ \\[3pt]
    {\istoy{t}:} Binary classification of whether an image contains a toy~$t \in \{ 0 \dots 4 \}$  \\[3pt]
    {\regressspeed:} Return the speed value and a maximum distance constant from an image of a speed limit sign. \\[3pt]
    {\regressmnist:} Return the value and a maximum distance constant from a digit image from MNIST dataset. \\[3pt]
    {\countdigit{d}:} Given a list of images, count the number of images of digit $d$ \\[3pt]
    {\counttoy{t}:} Given a list of images, count the number of images of toy $t$ \\[3pt]
    {\sumdigits:} Given a list of digit images, compute the sum of the digits. \\[3pt]
    {\shortestpathstreet:} Given a grid of images of speed limit signs, find the shortest distances to all
    other nodes  \\[3pt]
    {\shortestpathmnist:} Given a grid of MNIST images, and a source node, find the shortest distances to all
    other nodes in the grid.  \\[4pt]
\end{minipage}%
\hspace{1em}%
\begin{minipage}[t]{0.65\textwidth}
\small\raggedright
    \textbf{Task Sequences}\\[3pt]
    \emph{Counting} \\[2pt]
    \textbf{CS1:} Evaluate low-level transfer. \\ {\em Task 1}: \recogdigit{d1}; {\em Task 2}: \recogdigit{d2}; {\em Task 3}: \countdigit{d1}; {\em Task 4}: \countdigit{d2}  \\[2pt]
    \textbf{CS2:} Evaluate high-level transfer, and learning of perceptual tasks  from higher-level supervision. \\ {\em Task 1}: \recogdigit{d1}; {\em Task 2}: \countdigit{d1}; {\em Task 3}: \countdigit{d2}; {\em Task 4}: \recogdigit{d2}  \\[2pt]
    \textbf{CS3:} Evaluate high-level transfer of counting across different image domains. \\  {\em Task 1}: \recogdigit{d}; {\em Task 2}: \countdigit{d}; {\em Task 3}: \counttoy{t}; {\em Task 4}: \istoy{t}  \\[4pt]
    \emph{Summing} \\[2pt]
    \textbf{SS:} Demonstrate low-level transfer of a multi-class classifier as well as the advantage of functional methods like foldl in specific situations. \\[2pt] 
    {\em Task 1}: \classifydigit; {\em Task 2}: \sumdigits  \\[4pt]
    \emph{Single-Source Shortest Path} \\[2pt]
    \textbf{GS1:}  Learning of complex algorithms. \\[2pt] {\em Task 1}: \regressspeed; {\em Task 2}: \shortestpathstreet \\[2pt]
    \textbf{GS2:}  High-level transfer of complex
    algorithms. \\ {\em Task 1}: \regressmnist; {\em Task 2}: \shortestpathmnist; {\em Task 3:} \shortestpathstreet \\[2pt]
    \emph{Long sequence} {\bf LS}.\\[2pt]
    {\em Task 1}:\countdigit{d1}; {\em Task 2}: \counttoy{t1}; {\em Task 3}: \istoy{t2}; {\em Task 4}: \recogdigit{d2}; {\em Task 5}: \counttoy{t3}; {\em Task 6}: \countdigit{d3}; {\em Task 7}: \counttoy{t4}; {\em Task 8}: \recogdigit{d4}; {\em Task 9}: \countdigit{d5} 
    \end{minipage}
    \caption[]{Tasks and task sequences.}\label{fig:tasks}
\vspace{-0.2in}
\end{figure}

\textbf{Experimental setup.}
We allow three kinds of neural library modules: multi-layer perceptrons (MLPs), convolutional neural networks (CNNs) and recurrent neural networks (RNNs). We use two symbolic synthesis strategies: top-down
refinement and evolutionary. We use three types of baselines: (1) \emph{standalone networks}, which
do not do transfer learning, but simply train a new network (an RNN) for each
task in the sequence, starting from random weights; (2) a
traditional 
neural approach to \emph{low-level transfer} (LLT) that transfers all
weights learned in the previous task, except for the output layer that
is kept task-specific; 
and (3) a version of the {\em progressive neural
  networks} (PNNs)~\cite{rusu2016progressive} approach, which retains a pool of
pretrained models during training and learns lateral connections among
these models. Experiments were performed using a single-threaded implementation on a Linux system, with 8-core Intel E5-2620 v4 2.10GHz CPUs and TITAN X (Pascal) GPUs.


The architecture chosen for the standalone and LLT baselines
composes an MLP, an RNN, and a CNN, and matches the
structure of a high-performing program returned by \system, to
enable an apples-to-apples comparison. In PNNs, every task in a sequence
is associated with a network with the above architecture; lateral
connections between these networks are learned. Each sequence
involving digit classes $d$ and toy classes $t$ was instantiated five
times for random values of $d$ and $t$, and the results shown are
averaged over these instantiations. In the graph sequences, we ran the same sequences with different random seeds, and shared the regressors learned for the first tasks across the competing methods for a more reliable comparison. 
We do not compare against PNNs in this case, as it is nontrivial to extend them to work with graphs.
We evaluate the competing approaches on 2\%, 10\%, 20\%, 50\% and 100\% of the training data for all but the graph sequences, where we evaluate only on 100\%.
For classification tasks, we report error, while for the regression
tasks --- counting, summing, regress\_speed and shortest\_path --- we
report root mean-squared error (RMSE).  

\begin{figure*}
{\footnotesize
   \centering
    \begin{tabular}{lll}
        \textbf{~~Task} & \textbf{Top 3 programs} & \textbf{RMSE} \tabularnewline
        \hline
        \begin{tabular}{l}Task 1: \regressmnist \end{tabular} & 1. $\mathrm{nn\_gs2\_1} \circ \mathrm{nn\_gs2\_2}$ & {1.47} \tabularnewline
        \hline
        \multirow{3}{*}{\begin{tabular}{l}Task 2:\\ \shortestpathmnist\end{tabular}} & 1. $(\convg^{10}~(\mathrm{nn\_gs2\_3})) \circ (\mapg~(\mathrm{lib.nn\_gs2\_1} \circ \mathrm{lib.nn\_gs2\_2}))$  & {1.57} \tabularnewline
        & 2. $(\convg^{9}~(\mathrm{nn\_gs2\_4})) \circ (\mapg~(\mathrm{lib.nn\_gs2\_1} \circ \mathrm{lib.nn\_gs2\_2}))$  & {1.72}
        \tabularnewline
        & 3. $(\convg^{9}~(\mathrm{nn\_gs2\_5})) \circ (\mapg~(\mathrm{nn\_gs2\_6} \circ \mathrm{nn\_gs2\_7}))$  & {4.99} \tabularnewline
        \hline
        \multirow{3}{*}{\begin{tabular}{l}Task 3:\\ \shortestpathstreet \end{tabular}} & 1. $(\convg^{10}(\mathrm{lib.nn\_gs2\_3})) \circ (\mapg~(\mathrm{nn\_gs2\_8} \circ \mathrm{nn\_gs2\_9}))$  & {3.48} \tabularnewline
        & 2. $(\convg^{9}(\mathrm{lib.nn\_gs2\_3})) \circ (\mapg~(\mathrm{nn\_gs2\_10} \circ \mathrm{nn\_gs2\_11}))$  & {3.84} \tabularnewline
        & 3. $(\convg^{10}(\mathrm{lib.nn\_gs2\_3})) \circ (\mapg~(\mathrm{lib.nn\_gs2\_1} \circ \mathrm{lib.nn\_gs2\_2}))$  & {6.91}
        \tabularnewline
    \end{tabular}
}
\caption{Top 3 synthesized programs on Graph Sequence 2 (GS2). Here, $\mapg$ denotes a graph map of the appropriate type; $\convg^i$ denotes $i$ repeated applications of a graph convolution combinator (of the appropriate type).}\label{fig:GS2Progs-main}
\end{figure*}

\textbf{Results: Synthesized programs.}
\system  successfully synthesizes programs for each of the tasks in Figure~\ref{fig:tasks} within at most 22 minutes. 
We list in Figure~\ref{fig:GS2Progs-main} the top 3 programs for each task in the graph sequence GS2, and the corresponding RMSEs. 
Here, function names with prefix ``nn\_" denote fresh neural modules trained during the corresponding tasks. Terms with prefix ``lib." denote pretrained neural modules selected from the library. 
The synthesis times for Task 1, Task 2, and Task 3 are 0.35s,
1061s, 
and 
1337s, 
respectively.

As an illustration, consider the top program for Task 3: $(\convg^{10}~\mathrm{lib.nn\_gs2\_3}) \circ (\mapg~(\mathrm{nn\_gs2\_8} \circ \mathrm{nn\_gs2\_9}))$. Here, $\mapg$ takes as argument a function for processing the images of speed limits. Applied to the input graph, the map returns a graph $G$ in which each node contains a number associated with its corresponding image and information about the least cost of travel to the node. The kernel for the graph convolution combinator $\convg$ is a function $\mathrm{lib.nn\_gs2\_3}$, originally learned in Task 2, that implements the {\em relaxation} operation used in shortest-path algorithms. The convolution is applied repeatedly, just like in the Bellman-Ford shortest path algorithm.

In the SS sequence, 
the top program for Task 2 is: $(\foldl$~nn\_ss\_3 zeros(1)) $\circ$ $\mapl$(nn\_ss\_4 $\circ$ lib.nn\_ss\_2). Here, $\foldl$ denotes the fold operator applied to lists, and 
zeros(dim) is a function that returns a zero tensor of appropriate dimension. 
The program uses map to apply a previously learned CNN feature extractor (lib.nn\_ss\_2) and a learned transformation of said features into a 2D hidden state, to all images in the input list. It then uses fold with another function (nn\_ss\_3) to give the final sum. Our results, presented in the supplementary material, show that this program greatly outperforms the baselines, even in the setting where all of the training data is available. We believe that this is because the synthesizer has selected a program with fewer parameters than the baseline RNN.
In the results for the counting sequences (CS) and the long sequence (LS), the number of evaluated programs is restricted to 20, therefore $\foldl$ is not used within the synthesized programs. This allows us to evaluate the advantage of \system brought by its transfer capabilities, rather than its rich language.

\begin{figure*}[t!]
\hspace{-1em}\begin{tabular}{@{}c@{}c@{}c@{}}
Low-level transfer (CS1)
&
High-level transfer (CS2)
&
\parbox{1.8in}{\centering High-level transfer \\ across domains (CS3)}
\\
    \begin{subfigure}[t]{2in}
        \centering
        \includegraphics[width=1.8in]{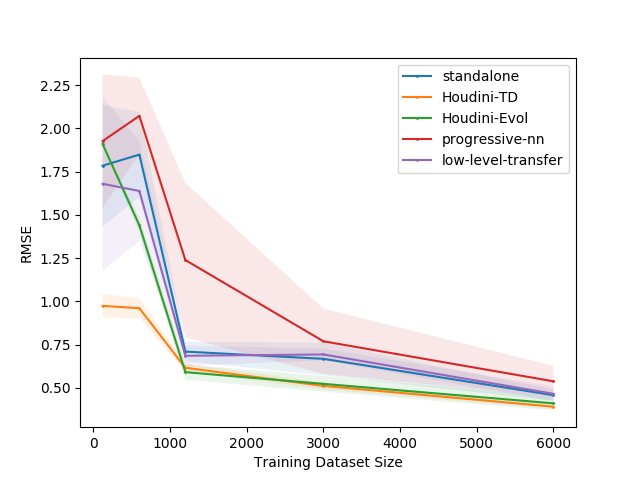}
        \caption{CS1 Task 3: \countdigit{d_1}}\label{fig:cs1t3}
    \end{subfigure}
&
    \begin{subfigure}[t]{2in}
        \centering
        \includegraphics[width=1.8in]{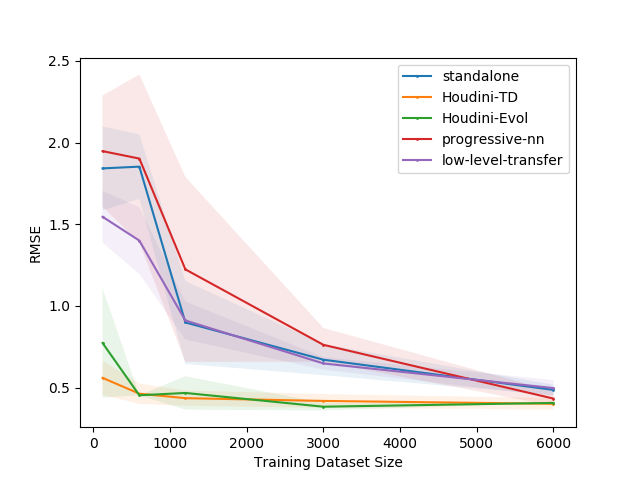}
        \caption{CS2 Task 3: \countdigit{d_2}}\label{fig:cs2t3}
    \end{subfigure}
&
    \begin{subfigure}[t]{2in}
        \centering
        \includegraphics[width=1.8in]{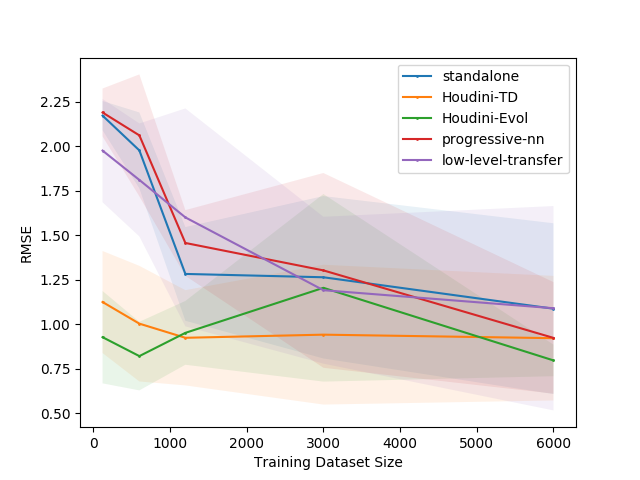}
        \caption{CS3 Task 3: \counttoy{t_1}}\label{fig:3t3}
    \end{subfigure}

\\    
    \begin{subfigure}[t]{2in}
        \centering
        \includegraphics[width=1.8in]{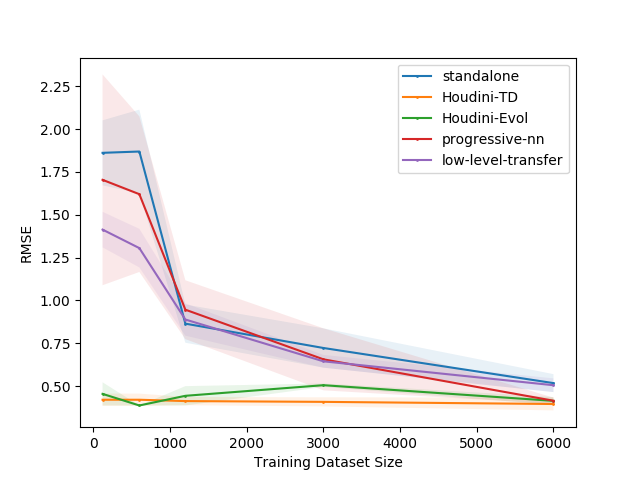}
        \caption{CS1 Task 4: \countdigit{d_2}}\label{fig:cs1t4}
    \end{subfigure}
&
    \begin{subfigure}[t]{2in}
        \centering
        \includegraphics[width=1.8in]{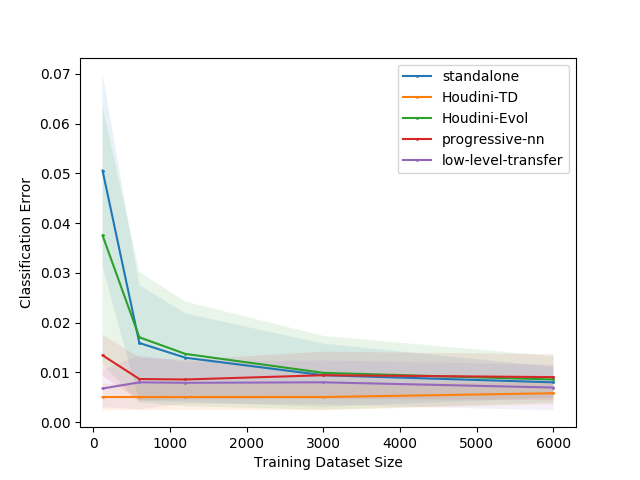}
        \caption{CS2 Task 4: \isdigit{d_2}}\label{fig:cs2t4}
    \end{subfigure}
&    \begin{subfigure}[t]{2in}
        \centering
        \includegraphics[width=1.8in]{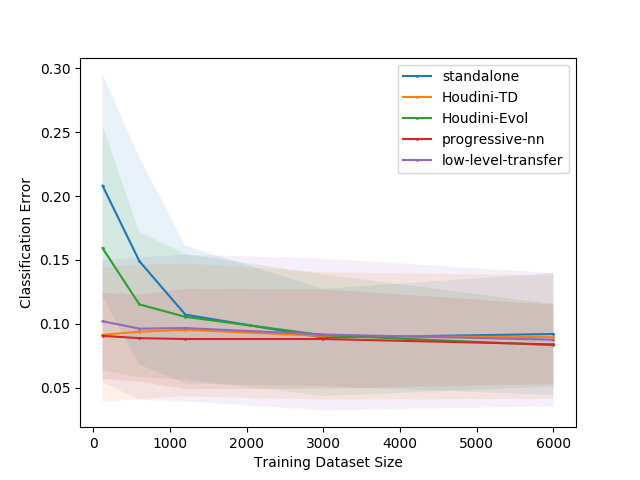}
        \caption{CS3 Task 4: \istoy{t_1}}\label{fig:3t4}
    \end{subfigure}
 \end{tabular}
    \caption{Lifelong ``learning to count'' (Sequences CS1 -- CS3), demonstrating
    both low-level transfer of perceptual concepts and high-level transfer
    of a counting network. {\sc Houdini-TD} and {\sc Houdini-Evol} are \system with the top-down and evolutionary synthesizers, respectively.}
    \label{fig:cs1}\label{fig:cs2}\label{fig:cs3}
\end{figure*}

\textbf{Results: Transfer.}
First we evaluate the performance of the methods on the counting 
sequences (Figure~\ref{fig:cs1}).  For space, we omit early tasks
where, by design, there is no opportunity for transfer; for these
results, see the Appendix.  In all cases where there is an opportunity
to transfer from previous tasks, we see that \system has much lower
error than any of the other transfer learning methods.  The actual
programs generated by \system are listed in the Appendix.


Task sequence CS1 evaluates the method's ability to 
selectively perform low-level transfer of a perceptual
concept across higher level tasks. 
The first task that provides a transfer opportunity is CS1 task 3 (Figure~\ref{fig:cs1t3}). There are two potential
lower-level tasks that the methods could transfer from:  \recogdigit{d_1} and \recogdigit{d_2}.
\system learns programs composed of neural modules $\mathrm{nn\_cs1\_1}$, $\mathrm{nn\_cs1\_2}$, $\mathrm{nn\_cs1\_3}$, and $\mathrm{nn\_cs1\_4}$
for these two tasks. During training for the \countdigit{d1} task, all the previously learned neural modules are available in the library. The learner, however, picks the correct module ($\mathrm{nn\_cs1\_2}$) for reuse, learning the program 
``$\mathrm{nn\_cs1\_7} \circ (\mapl~(\mathrm{nn\_cs1\_8} \circ \mathrm{lib.nn\_cs1\_2}))$''
where $\mathrm{nn\_cs1\_7}$ and  $\mathrm{nn\_cs1\_8}$ are fresh neural modules, and $\mapl$ stands for a list map combinator of appropriate type.
The low-level transfer baseline cannot select which of the previous tasks to re-use,
and so suffers worse performance.

Task sequence CS2 provides
an opportunity to transfer the higher-level concepts 
 of counting, across different digit classification tasks.
Here CS2 task 3 (Figure~\ref{fig:cs2t3}) is the task that provides the first opportunity for transfer.
We see that \system is able to learn much faster on this task because it is able to reuse a network
which has learned from the previous counting task.
%
Task sequence CS3 examines whether the methods can demonstrate high-level transfer
when the input image domains are very different, from the MNIST domain to the NORB domain
of toy images. We see in Figure~\ref{fig:3t3} that the
higher-level network still successfully transfers across tasks, learning an effective network
for counting the number of toys of type $t_1$, even though the network has
not previously seen any toy images at all. What is more, it can be seen that because
of the high-level transfer, \system has learned a modular solution to this problem.
From the subsequent performance on a standalone toy classification task (Figure~\ref{fig:3t4}),
we see that CS3 task 3 has already caused
the network to induce a re-usable classifier on toys.
Overall, it can be seen that \system outperforms all the baselines even under the limited data setting, confirming the successful selective transfer of both low-level and high-level perceptual information.
Similar results can be seen on the summing task (see supplementary material).
Moreover, on the longer task sequence LS, we also find that \system performs significantly better
on the tasks in the sequence where there is an opportunity for transfer, and performs comparably
the baselines on the other tasks (see supplementary material). Furthermore, on the summing sequence, our results also show low level transfer.

Finally, for the graph-based tasks (Table~\ref{gs1}), we see that the graph convolutional
program learned by \system on the graph tasks has significantly less error than a simple sequence model, a standalone baseline and the evolutionary-algorithm-based version of \system.
As explained earlier, in the \shortestpathstreet \ task in the graph sequence GS2, \system learns a program
that uses newly learned regress functions for
the street signs, along with a ``relaxation"
function already learned from the earlier task \shortestpathmnist.
In Table~\ref{gs2}, we see 
this program performs well, 
suggesting that a domain-general relaxation operation is being learned. Our approach also outperforms the low-level-transfer baseline, except on the \shortestpathstreet \ task in GS2. 
We are unable to compare directly to NTPT because no public implementation is available.
However, our graph task is a more difficult version of a task from \citet{GauntBKT17}, 
who report on their shortest-path task ``2\% of random restarts successfully converge to a program that generalizes''
(see their supplementary material).

\textbf{Results: Typed vs. untyped synthesis.}
To assess the impact of our type system, we count the programs that \symmod produces with and without a type system (we pick the top-down implementation for this test, but the results also apply to the evolutionary synthesizer). 
Let the {\em size} of a program be the number of occurrences of library functions and combinators in the program. 
Table~\ref{tab:types} shows the number of programs of different sizes generated for the tasks in the sequence CS1. 
Since the typical program size in our sequences is less than 6, we vary the target program size from
4 to 6. When the type system is disabled, the only constraint that
\symmod has while composing programs is the arity of the library functions.
We note that this constraint fails to bring down the number of
candidate programs to a manageable size. With the type system,
however, \symmod produces far fewer candidate programs.
For reference, neural architecture search
often considers thousands of potential architectures 
for a single task \cite{liu17}.

\textbf{Results: Top-Down vs. Evolutionary Synthesis.}

Overall, the top-down implementation of \symmod outperformed the evolutionary implementation. In some tasks, the two strategies performed about the same. However, the evolutionary strategy has high variance; indeed, in many runs of the task sequences, it times out without finding a solution. The timed out runs are not included in the plots.

\begin{table}
\begin{tabular}{ll}
\begin{minipage}{3in}
{\footnotesize 
    \begin{tabular}{lcrrr}
        \hline
        \multirow{2}{*}{} & \multirow{2}{*}{\textbf{Task}} & \multicolumn{3}{c}{\textbf{Number of programs}}\tabularnewline
        \cline{3-5}
        & & size = 4 & size = 5 & size = 6\tabularnewline
        \hline
        \multirow{4}{*}{No types} & Task 1 & 8182 & 110372 & 1318972\tabularnewline
        & Task 2 & 12333 & 179049 & 2278113\tabularnewline
        & Task 3 & 17834 & 278318 & 3727358\tabularnewline
        & Task 4 & 24182 & 422619 & 6474938\tabularnewline
        \hline
        \multirow{4}{*}{+ Types} & Task 1 & 2 & 20 & 44\tabularnewline
        & Task 2 & 5 & 37 & 67\tabularnewline
        & Task 3 & 9 & 47 & 158\tabularnewline
        & Task 4 & 9 & 51 & 175\tabularnewline
        \hline
    \end{tabular}
}
    \caption{Effect of the type system on the number of programs considered 
    in the symbolic search for task sequence CS1.}
    \label{tab:types}
\end{minipage}
 &
 \begin{minipage}{2.5in}
    \centering
    \begin{tabular}{rcc}
        & Task 1 & Task 2 \\
        RNN w llt & 0.75 & 5.58  \\
        standalone & 0.75 & 4.96  \\
        \system & 0.75 & 1.77 \\
        \system EA & 0.75 & 8.32 \\
        low-level-transfer & 0.75 & 1.98
    \end{tabular}
    
    {\small (a) Low-level transfer (llt)  (task sequence GS1).}
     \\[1ex]

    \begin{tabular}{rccc}
        & Task 1 & Task 2 & Task 3 \\
        \normalsize
        RNN w llt & 1.44 & 5.00 & 6.05 \\
        standalone & 1.44 & 6.49 & 7.  \\
        \system & 1.44 & 1.50 & 3.31 \\
        \system EA & 1.44 & 6.67 & 7.88 \\
        low-level-transfer & 1.44 & 1.76 & 2.08
    \end{tabular}

    {\small (b) High-level transfer (task sequence GS2).}
    \\[1ex]
    
\captionof{table}{Lifelong learning on graphs.
Column 1: RMSE on speed/distance from image. Columns 2, 3: RMSE on shortest path (mnist, street).}\label{gs1}\label{gs2}
\end{minipage}
\end{tabular}
\end{table}

    \section{Conclusion}\seclabel{conc}

    We have presented \system, the first neurosymbolic approach to the synthesis of differentiable functional programs. Deep networks can be naturally
specified as differentiable programs, and functional programs can compactly
represent popular deep architectures \cite{olahblog}.
Therefore,  symbolic search through a space of differentiable functional
programs is particularly appealing, because it can at the same time select both which pretrained neural library functions
should be reused, and also what deep architecture should be used to combine them. 
On several lifelong learning tasks that combine perceptual and algorithmic
reasoning, we showed that \system can accelerate learning by transferring high-level concepts.

    \bibliographystyle{plain}

    \bibliography{main}
\newpage

\appendix

\section{Assigning Types to \system Programs}\label{app:types}

A program $e$ in \system is assigned a type using the following rules:
\begin{itemize}
\item  $e = e' \circ e''$ is assigned a type iff $e'$ has type $\tau \mto \tau'$ and $e''$ has type
  $\tau' \mto \tau''$. In this case, $e$ has type $\tau \mto
  \tau''$. 
\item $e = \mapc_{\alpha\zug{\tau}}~e'$ is assigned a type
  iff $e'$ has the type $\tau \mto \tau'$. In this case, the type of $e$
  is $\alpha\zug{\tau} \mto \alpha\zug{\tau'}$.
\item $e = \foldc_{\alpha\zug{\tau}}~e'~z$ is assigned a type
  iff $e'$ has the type $\tau' \mto (\tau \mto \tau')$ and $z$ has the type $\tau'$. In this case,
  $e$ has type
  $\alpha\zug{\tau} \mto \tau'$.
\item $e = \convc_{\alpha\zug{\tau}}~e'$ is assigned a type
  iff $e'$ has the type $\mathtt{list}\zug{\tau} \mto \tau'$. In this case, $e$
  has type $\alpha\zug{\tau} \mto \alpha\zug{\tau'}$.
\end{itemize}
If it is not possible to assign a type to the program $e$,
then it is considered {\em type-inconsistent} and excluded from the
scope of synthesis.

\section{Symbolic Program Synthesis}\label{app:synthesis}

\newcommand{\pe}{\mathit{pe}}

In this appendix we provide implementation details of our
synthesis algorithms.

\subsection{Synthesis Using Top-down Iterative Refinement}

Now we give more details on the implementation of \symmod based on
iterative refinement.  To explain this algorithm, we need to define a
notion of a {\em partial program}. The grammar for partial programs
$e$ is obtained by augmenting the \system grammar (\figref{language})
with an additional rule: $e ::= \hole{\tau}$. The form $\hole{\tau}$
represents a {\em hole}, standing for missing code. A program with
holes has no operational meaning; however, we do have a type system
for such programs. This type system follows the rules in
Appendix~\ref{app:types}, but in addition, axiomatically assumes any subterm $\hole{\tau}$
to be of type $\tau$. A partial program that cannot be assigned a type
is automatically excluded from the scope of synthesis.

Now, the initial input to the algorithm is the type $\tau$ of the function
we want to learn. The procedure proceeds iteratively, maintaining a
priority queue $Q$ of {\em synthesis subtasks} of the form $(e, f)$,
where $e$ is a type-safe partial or complete program of type $\tau$,
and $f$ is either a hole of type $\tau'$ in $e$, or a special symbol
$\perp$ indicating that $e$ is complete (i.e., free of holes). The interpretation of such a
task is to find a replacement $e'$ of type $\tau'$ for the hole $f$
such that the program $e''$ obtained by substituting $f$ by $e'$ is
complete. (Because $e$ is type-safe by construction, $e''$ is of type
$\tau$.)  The queue is sorted according to a heuristic cost function
that prioritizes simpler programs.

Initially, $Q$ has a single element $(e, f)$, where $e$ is an ``empty''
program of form $\hole{\tau}$, and $f$ is
a reference to the hole in $e$. The procedure iteratively processes subtasks in the queue $Q$,
selecting a task $(e, f)$ in the beginning of each iteration. If the
program $e$ is complete, it is sent to the neural module for parameter
learning. Otherwise, the algorithm expands the program $e$ by
proposing a partial program that fills the hole $f$. To do this, the
algorithm selects a production rule for partial programs from the
grammar for partial programs. Suppose the right hand side of this rule
is $\alpha$. The algorithm constructs an expression $e'$ from $\alpha$
by replacing each nonterminal in $\alpha$ by a hole with the same type
as the nonterminal. If $e'$ is not of the same type as $f$, it is
automatically rejected. Otherwise, the algorithm constructs the
program $e'' = e[f \mapsto e']$. For each hole $f'$ in $e''$, the
algorithm adds to $Q$ a new task $(e'', f')$. If $e''$ has no hole, it
adds to $Q$ a task $(e'', \perp)$.

\subsection{Evolutionary Synthesis}

The evolutionary synthesis algorithm is an iterative procedure that maintains a population of
programs. The population is initialized with a set of randomly
generated type-safe parameterized programs. Each iteration of the algorithm performs the following steps. 
\begin{enumerate}
\item Each program in the population is sent to the neural module
  \neuromod, which computes a {\em fitness score} (the loss under optimal
  parameters) for the program.

\item  We perform {\em random proportional selection}, in which a subset
  of the (parameterized) programs are retained, while the other programs are filtered
  out. Programs with higher fitness are more likely to remain in the
  population. 

\item We perform a series of {\em crossover} operations, each of which
  draws a random pair of programs from the population and swaps a pair
  of randomly drawn subterms of the same type in these programs.

\item We perform a series of {\em mutation} operations, each of which randomly
  chooses a program and replaces a random
  subterm the program with a new subterm of the same type.
\end{enumerate}

Because the crossover and mutation operations only replace terms with
other terms of the same type, the programs in the population are
always guaranteed to be type-consistent. This fact is key to the
performance of the algorithm.

\section{Details of Experimental Setup}

The initial library models, which have trainable weights, have the following architecture. MLP modules have one hidden layer of size 1024, followed by batch normalization and dropout, followed by an output layer. CNNs have two convolutional layers with 32 and 64 output channels respectively, each with a 5x5 kernel, stride 1 and 0 padding, and each followed by max pooling, followed by spatial dropout. RNN modules are long short-term memory (LSTM) networks with a hidden dimension of 100, followed by an output layer, which transforms the last hidden state. For a given task, we use the input and output types of the new function to decide between MLP, CNN, or RNN, and also deduce the output activation function.

The standalone baseline for counting uses an architecture of the form $\lambda x. \mathrm{RNN}(\mapc~(\mathrm{MLP} \circ \mathrm{CNN}(x)))$, which is intuitively appropriate for the task, and also matches the shape of some programs commonly returned by \system. 
 
As for the shortest path sequences, 
the first task for GS1 and GS2 is regression, which we train using a network with architecture $\mathrm{MLP} \circ \mathrm{CNN}$, in which the last layer is linear.
In the RNN baseline for the other tasks in the graph sequences, we map a learned $\mathrm{MLP} \circ \mathrm{CNN}$ regression module to each image in the grid. Afterwards, we linearize the grid row-wise, converting it into a list, and then we process it using an LSTM (RNN) with hidden state of size 100. The number was chosen so that both our implementation and the baseline have almost the same number of parameters.

For multi-class classification (Sequence SS - Task 1) and regression (GS1 - Task1, GS2 - Task 1), we used all training images available. For the rest of the tasks in GS1, GS2, GS3 and SS, we use 12000 data points for training, with 2100 for testing. The list lengths for training are [2, 3, 4, 5], and [6, 7, 8] for testing in order to evaluate the generalization to longer sequences. We train for 20 epochs on all list-related tasks and for 1000 epochs for the regression tasks. The training datasets for the graph shortest path tasks (GS1 - Task 2; GS2 - Task2, GS2 - Task3) consists of 70,000 3x3 grids and 1,000,000 4x4 grids, while the testing datasets consists of 10,000 5x5 grids. The number of epochs for these tasks is 5.
In GS2 - Task3, the \emph{low-level transfer} baseline reuses the regression function learned in GS2 - Task1, thus, the image dimensions from MNIST and the colored GTSRB need to match. Therefore, we expanded the MNIST digit images, used for the graph sequences GS1 and GS2, to 28x28x3 dimensionality and resized the images from GTSRB from 32x32x3 to 28x28x3.

For all experiments, we use early stopping, reporting the the test error
at the epoch where the validation error was minimized.

\section{Programs Discovered in Experiments}\label{app:programs}


Tables \ref{tab:CS1Progs}-\ref{tab:LS5ProgsEvo} list the top 3 programs and the corresponding classification errors/RMSEs, on a test dataset, for most of our task sequences. The programs are ordered by their performance on a validaiton dataset. Finally, the presented programs are the ones evaluated for all (100\%) of the training dataset.
Here we use the syntax $\composec$ to denote function composition. Program terms with prefix ``nn\_" denote neural modules trained during the corresponding tasks whereas terms with prefix ``lib." denote already trained neural modules in the library.
For example, in Counting Sequence 1 (Table \ref{tab:CS1Progs}), ``nn\_cs1\_1" is the neural module trained during Task 1 (\recogdigit{d_1}).
After completion of this task, the neural module is added to the library and is available for use during the subsequent tasks.
For example, the top performing program for Task 3 (\countdigit{d_1})
uses the neural module ``lib.nn\_cs1\_1" from the library (and a
freshly trained neural module ``nn\_cs1\_5") to construct a program
for the counting task.

\section{Summing Experiment}\label{app:summing}

In this section we present the result
from task sequence SS in Figure 3 of the main paper.
This sequence was designed to demonstrate
low-level transfer of a multi-class classifier as well as the advantage of functional methods like foldl in specific situations.
The first task of the sequence is a simple MNIST
classifier, on which all competing methods do equally
well. The second task is a regression task, to learn
to sum all of the digits in the sequence. The standalone method,
low level transfer one and the progressive neural networks all perform equally poorly (note that their lines are overplotted in the Figure),
but the synthesized program from \system
is able to learn this function easily
because it is able to use a foldl operation. We also add a new baseline "standalone\_with\_fold", which reuses the program found by \system, but trains the parameter from a random initialization.

\begin{figure}[h!]
    \centering
    \begin{subfigure}[t]{3in}
        \centering
        \includegraphics[width=3in]{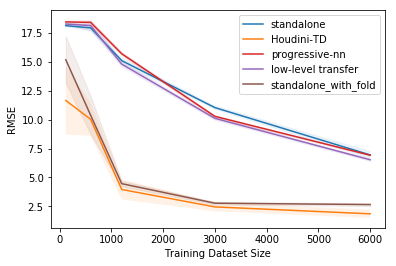}
        \caption{Task 2: Sum digits}\label{fig:4t2} \label{fig:sst2}
    \end{subfigure}
    \caption{Lifelong learning for ``learning to sum'' (Sequence SS).}\label{fig:ss}
\end{figure}

\begin{table*}[]
{\small
\centering
\scriptsize
\caption{Counting Sequence 1(CS1). ``CE'' denotes classification error and ``RMSE''
denotes root mean square error. }
\label{tab:CS1Progs}
\begin{tabular}{l>{\raggedright}p{\progcolwidth}r}
    \textbf{Task} & \textbf{Top 3 programs} & \score \tabularnewline
    \hline
    Task 1: \recogdigit{d_1} & 1. $\composec$(nn\_cs1\_1, nn\_cs1\_2) & \error{1} \tabularnewline
    \hline
    \multirow{3}{*}{Task 2: \recogdigit{d_2}} & 1. $\composec$(nn\_cs1\_3, nn\_cs1\_4) & \error{1} \tabularnewline
    & 2. $\composec$(nn\_cs1\_5, lib.nn\_cs1\_2) & \error{1} \tabularnewline
    & 3. $\composec$(lib.nn\_cs1\_1, nn\_cs1\_6) & \error{1} \tabularnewline
    \hline
    \multirow{3}{*}{Task 3: \countdigit{d_1}} & 1. $\composec$(nn\_cs1\_7, $\mapl$($\composec$(nn\_cs1\_8, lib.nn\_cs1\_2))) & \rmse{0.38} \tabularnewline
    & 2. $\composec$($\composec$(nn\_cs1\_9, $\mapl$(nn\_cs1\_10)), $\mapl$(nn\_cs1\_11)) & \rmse{0.38} \tabularnewline
    & 3. $\composec$(nn\_cs1\_12, $\composec$($\mapl$(nn\_cs1\_13), $\mapl$(lib.nn\_cs1\_2))) & \rmse{0.40} \tabularnewline
    \hline
    \multirow{3}{*}{Task 4: \countdigit{d_2}} & 1. $\composec$(nn\_cs1\_14, $\composec$($\mapl$(lib.nn\_cs1\_1),
    $\mapl$(nn\_cs1\_15))) & \rmse{0.32} \tabularnewline
    & 2. $\composec$(lib.nn\_cs1\_7, $\mapl$($\composec$(nn\_cs1\_16, lib.nn\_cs1\_4))) & \rmse{0.37} \tabularnewline
    & 3. $\composec$(lib.nn\_cs1\_7, $\composec$($\mapl$(nn\_cs1\_17), $\mapl$(lib.nn\_cs1\_4))) & \rmse{0.37}
    \tabularnewline
\end{tabular}
}
\end{table*}

\begin{table*}[]
{\small
\centering
\scriptsize
\caption{Counting Sequence 2(CS2)}
\label{tab:CS2Progs}
\begin{tabular}{l>{\raggedright}p{\progcolwidth}r}
    \textbf{Task} & \textbf{Top 3 programs} & \score \tabularnewline
    \hline
    Task 1: \recogdigit{d_1} & 1. $\composec$(nn\_cs2\_1, nn\_cs2\_2) & \error{1} \tabularnewline
    \hline
    \multirow{2}{*}{Task 2: \countdigit{d_1}} & 1. $\composec$(nn\_cs2\_3, $\composec$($\mapl$(nn\_cs2\_4), $\mapl$(nn\_cs2\_5))) & \rmse{0.35} \tabularnewline
    & 2. $\composec$(nn\_cs2\_6, $\mapl$($\composec$(nn\_cs2\_7, nn\_cs2\_8))) & \rmse{0.40} \tabularnewline
    & 3. $\composec$($\composec$(nn\_cs2\_9, $\mapl$(nn\_cs2\_10)), $\mapl$(nn\_cs2\_11)) &  \rmse{0.41} \tabularnewline
    \hline
    \multirow{3}{*}{Task 3: \countdigit{d_2}} & 1. $\composec$(lib.nn\_cs2\_3, $\composec$($\mapl$(nn\_cs2\_12), $\mapl$(lib.nn\_cs2\_2))) & \rmse{0.34} \tabularnewline
    & 2. $\composec$(lib.nn\_cs2\_3, $\mapl$($\composec$(nn\_cs2\_13, lib.nn\_cs2\_2))) & \rmse{0.33} \tabularnewline
    & 3. $\composec$(lib.nn\_cs2\_3, $\composec$($\mapl$(nn\_cs2\_14), $\mapl$(nn\_cs2\_15))) & \rmse{0.33} \tabularnewline
    \hline
    \multirow{3}{*}{Task 4: \recogdigit{d_2}} & 1. $\composec$(nn\_cs2\_16, nn\_cs2\_17) & \error{1} \tabularnewline
    & 2. $\composec$(nn\_cs2\_18, lib.nn\_cs2\_2) &  \error{1} \tabularnewline
    & 3. $\composec$(lib.nn\_cs2\_12, lib.nn\_cs2\_2) & \error{1} \tabularnewline
\end{tabular}
}
\end{table*}

\begin{table*}[]
{\small
\centering
\scriptsize
\caption{Counting Sequence 3(CS3)}
\label{tab:CS3Progs}
\begin{tabular}{l>{\raggedright}p{\progcolwidth}r}
    \textbf{Task} & \textbf{Top 3 Programs} & \score \tabularnewline
    \hline
    Task 1: \recogdigit{d} & 1. $\composec$(nn\_cs3\_1, nn\_cs3\_2) & \error{1} \tabularnewline
    \hline
    \multirow{2}{*}{Task 2: \countdigit{d}} & 1. $\composec$(nn\_cs3\_3, $\mapl$($\composec$(nn\_cs3\_4, nn\_cs3\_5)))  & \rmse{0.40} \tabularnewline
    & 2. $\composec$(nn\_cs3\_6, $\mapl$($\composec$(nn\_cs3\_7, lib.nn\_cs3\_2)))  & \rmse{0.40} \tabularnewline
    & 3. $\composec$(nn\_cs3\_8, $\composec$($\mapl$(nn\_cs3\_9), $\mapl$(lib.nn\_cs3\_2)))  & \rmse{0.41} \tabularnewline
    \hline
    \multirow{3}{*}{Task 3: \counttoy{t}} & 1. $\composec$(lib.nn\_cs3\_3, $\composec$($\mapl$(nn\_cs3\_10), $\mapl$(nn\_cs3\_11))) & -0.73 \tabularnewline
    & 2. $\composec$(lib.nn\_cs3\_3, $\mapl$($\composec$(nn\_cs3\_12, nn\_cs3\_13))) & \rmse{0.67} \tabularnewline
    & 3. $\composec$(lib.nn\_cs3\_3, $\composec$($\mapl$(lib.nn\_cs3\_1), $\mapl$(nn\_cs3\_14))) & \rmse{0.96} \tabularnewline
    \hline
    \multirow{3}{*}{Task 4: \istoy{t}} & 1. $\composec$(nn\_cs3\_15, lib.nn\_cs3\_11) & \error{7} \tabularnewline
    & 2. $\composec$(nn\_cs3\_16, nn\_cs3\_17) & \error{5} \tabularnewline
    & 3. $\composec$(lib.nn\_cs3\_10, lib.nn\_cs3\_11) & \error{6} \tabularnewline
\end{tabular}
}
\end{table*}

\begin{table*}[]
{\small
\centering
\scriptsize
\caption{Summing Sequence(SS)}
\label{tab:SSProgs}
\begin{tabular}{l>{\raggedright}p{\progcolwidth}r}
    \textbf{Task} & \textbf{Top 3 programs} & \score \tabularnewline
    \hline
    Task 1: \classifydigit & 1. $\composec$(nn\_ss\_1, nn\_ss\_2) & \error{1}\tabularnewline
    \hline
    \multirow{2}{*}{Task 2: \sumdigits} & 1. $\composec$(($\foldl$ nn\_ss\_3 zeros(1)), $\mapl$($\composec$(nn\_ss\_4, lib.nn\_ss\_2)))
    & \rmse{2.15} \tabularnewline
    & 2. $\composec$(($\foldl$ nn\_ss\_5 zeros(1)), $\mapl$($\composec$(nn\_ss\_6, nn\_ss\_7))) & \rmse{2.58} \tabularnewline
    & 3. $\composec$(($\foldl$ nn\_ss\_8 zeros(1)), $\mapl$($\composec$(lib.nn\_ss\_1, lib.nn\_ss\_2))) & \rmse{4.30} \tabularnewline
\end{tabular}
}
\end{table*}

\begin{table*}[]{\small \centering \scriptsize
  \caption{Graph Sequence 1(GS1)}
  \label{tab:GS1Progs}
  \begin{tabular}{l>{\raggedright}p{\progcolwidth}r}
    \textbf{Task} & \textbf{Top 3 Programs} & \score \tabularnewline
\hline
\multirow{1}{*}{Task 1: \regressspeed}
& 1. $\composec$(nn\_gs1\_1, nn\_gs1\_2) & \rmse{0.64} \tabularnewline
\hline
\multirow{3}{*}{Task 2: \shortestpathstreet}
& 1. $\composec$($\convg^{10}$(nn\_gs1\_3), $\mapg$($\composec$(lib.nn\_gs1\_1, lib.nn\_gs1\_2))) & \rmse{1.88} \tabularnewline
& 2. $\composec$($\convg^{9}$(nn\_gs1\_4), $\mapg$($\composec$(lib.nn\_gs1\_1, lib.nn\_gs1\_2))) & \rmse{2.02} \tabularnewline
& 3. $\composec$($\convg^{10}$(nn\_gs1\_5), $\mapg$($\composec$(nn\_gs1\_6, nn\_gs1\_7))) & \rmse{6.76} \tabularnewline
   \end{tabular}
}\end{table*}


\begin{table*}[]{\small \centering \scriptsize
  \caption{Graph Sequence 2(GS2)}
  \label{tab:GS2Progs}
  \begin{tabular}{l>{\raggedright}p{\progcolwidth}r}
    \textbf{Task} & \textbf{Top 3 Programs} & \score \tabularnewline
\hline
\multirow{1}{*}{Task 1: \regressmnist}
& 1. $\composec$(nn\_gs2\_1, nn\_gs2\_2) & \rmse{1.47} \tabularnewline
\hline
\multirow{3}{*}{Task 2: \shortestpathmnist}
& 1. $\composec$($\convg^{10}$(nn\_gs2\_3), $\mapg$($\composec$(lib.nn\_gs2\_1, lib.nn\_gs2\_2))) & \rmse{1.57} \tabularnewline
& 2. $\composec$($\convg^{9}$(nn\_gs2\_4), $\mapg$($\composec$(lib.nn\_gs2\_1, lib.nn\_gs2\_2))) & \rmse{1.73} \tabularnewline
& 3. $\composec$($\convg^{9}$(nn\_gs2\_5), $\mapg$($\composec$(nn\_gs2\_6, nn\_gs2\_7))) & \rmse{4.99} \tabularnewline
\hline
\multirow{3}{*}{Task 3: \shortestpathstreet}
& 1. $\composec$($\convg^{10}$(lib.nn\_gs2\_3), $\mapg$($\composec$(nn\_gs2\_8, nn\_gs2\_9))) & \rmse{3.48} \tabularnewline
& 2. $\composec$($\convg^{9}$(lib.nn\_gs2\_3), $\mapg$($\composec$(nn\_gs2\_10, nn\_gs2\_11))) & \rmse{3.84} \tabularnewline
& 3. $\composec$($\convg^{10}$(lib.nn\_gs2\_3), $\mapg$($\composec$(lib.nn\_gs2\_1, lib.nn\_gs2\_2))) & \rmse{6.92} \tabularnewline
   \end{tabular}
}\end{table*}


\begin{table*}[]{\small \centering \scriptsize
\caption{Long Sequence 1(LS1).}
\label{tab:LS1Progs}
\begin{tabular}{l>{\raggedright}p{\progcolwidth}r}
    \textbf{Task} & \textbf{Top 3 Programs} & \score \tabularnewline
    \hline
    \multirow{3}{*}{Task 1: \countdigit{7}}
    & 1. $\composec$($\composec$(nn\_ls1\_1, $\mapl$(nn\_ls1\_2)), $\mapl$(nn\_ls1\_3)) & \rmse{0.46} \tabularnewline
    & 2. $\composec$(nn\_ls1\_4, $\mapl$($\composec$(nn\_ls1\_5, nn\_ls1\_6))) & \rmse{0.49} \tabularnewline
    & 3. $\composec$(nn\_ls1\_7, $\composec$($\mapl$(nn\_ls1\_8), $\mapl$(nn\_ls1\_9))) & \rmse{0.51} \tabularnewline
    \hline
    \multirow{3}{*}{Task 2: \countdigit{4}}
    & 1. $\composec$(lib.nn\_ls1\_1, $\composec$($\mapl$(nn\_ls1\_10), $\mapl$(nn\_ls1\_11))) & \rmse{1.50} \tabularnewline
    & 2. $\composec$(lib.nn\_ls1\_1, $\mapl$($\composec$(nn\_ls1\_12, nn\_ls1\_13))) & \rmse{1.61} \tabularnewline
    & 3. $\composec$($\composec$(lib.nn\_ls1\_1, $\mapl$(nn\_ls1\_14)), $\mapl$(nn\_ls1\_15)) & \rmse{1.64} \tabularnewline
    \hline
    \multirow{3}{*}{Task 3: \istoy{0}}
    & 1. $\composec$(nn\_ls1\_16, lib.nn\_ls1\_11) & \error{9.81} \tabularnewline
    & 2. $\composec$(nn\_ls1\_17, nn\_ls1\_18) & \error{8.86} \tabularnewline
    & 3. $\composec$(nn\_ls1\_19, lib.nn\_ls1\_3) & \error{12.86} \tabularnewline
    \hline
    \multirow{3}{*}{Task 4: \recogdigit{9}}
    & 1. $\composec$(nn\_ls1\_20, nn\_ls1\_21) & \error{1.38} \tabularnewline
    & 2. $\composec$(nn\_ls1\_22, lib.nn\_ls1\_3) & \error{2.14} \tabularnewline
    & 3. $\composec$(lib.nn\_ls1\_2, nn\_ls1\_23) & \error{1.95} \tabularnewline
    \hline
    \multirow{3}{*}{Task 5: \countdigit{2}}
    & 1. $\composec$(nn\_ls1\_24, $\composec$($\mapl$(nn\_ls1\_25), $\mapl$(nn\_ls1\_26))) & \rmse{1.08} \tabularnewline
    & 2. $\composec$(lib.nn\_ls1\_1, $\mapl$($\composec$(nn\_ls1\_27, nn\_ls1\_28))) & \rmse{1.02} \tabularnewline
    & 3. $\composec$(lib.nn\_ls1\_1, $\mapl$($\composec$(lib.nn\_ls1\_16, nn\_ls1\_29))) & \rmse{0.95} \tabularnewline
    \hline
    \multirow{3}{*}{Task 6: \countdigit{9}}
    & 1. $\composec$(lib.nn\_ls1\_1, $\mapl$($\composec$(nn\_ls1\_30, nn\_ls1\_31))) & \rmse{0.49} \tabularnewline
    & 2. $\composec$(lib.nn\_ls1\_1, $\mapl$($\composec$(nn\_ls1\_32, lib.nn\_ls1\_21))) & \rmse{0.49} \tabularnewline
    & 3. $\composec$(lib.nn\_ls1\_1, $\mapl$($\composec$(nn\_ls1\_33, lib.nn\_ls1\_3))) & \rmse{0.49} \tabularnewline
    \hline
    \multirow{3}{*}{Task 7: \countdigit{0}}
    & 1. $\composec$(nn\_ls1\_34, $\mapl$($\composec$(lib.nn\_ls1\_16, lib.nn\_ls1\_11))) & \rmse{0.94} \tabularnewline
    & 2. $\composec$(lib.nn\_ls1\_1, $\mapl$($\composec$(nn\_ls1\_35, nn\_ls1\_36))) & \rmse{0.81} \tabularnewline
    & 3. $\composec$(nn\_ls1\_37, $\mapl$($\composec$(nn\_ls1\_38, nn\_ls1\_39))) & \rmse{0.85} \tabularnewline
    \hline
    \multirow{3}{*}{Task 8: \recogdigit{7}}
    & 1. $\composec$(lib.nn\_ls1\_2, lib.nn\_ls1\_3) & \error{0.86} \tabularnewline
    & 2. $\composec$(nn\_ls1\_40, lib.nn\_ls1\_3) & \error{1.19} \tabularnewline
    & 3. $\composec$(nn\_ls1\_41, lib.nn\_ls1\_21) & \error{1.05} \tabularnewline
    \hline
    \multirow{3}{*}{Task 9: \countdigit{2}}
    & 1. $\composec$(nn\_ls1\_42, $\mapl$($\composec$(nn\_ls1\_43, lib.nn\_ls1\_26))) & \rmse{0.43} \tabularnewline
    & 2. $\composec$(lib.nn\_ls1\_1, $\mapl$($\composec$(nn\_ls1\_44, nn\_ls1\_45))) & \rmse{0.45} \tabularnewline
    & 3. $\composec$(lib.nn\_ls1\_1, $\mapl$($\composec$(nn\_ls1\_46, lib.nn\_ls1\_3))) & \rmse{0.45} \tabularnewline
\end{tabular}
}\end{table*}

\begin{table*}[]{\small \centering \scriptsize
\caption{Long Sequence 2(LS2).}
\label{tab:LS2Progs}
\begin{tabular}{l>{\raggedright}p{\progcolwidth}r}
    \textbf{Task} & \textbf{Top 3 Programs} & \score \tabularnewline
    \hline
    \multirow{3}{*}{Task 1: \countdigit{1}}
    & 1. $\composec$(nn\_ls2\_1, $\mapl$($\composec$(nn\_ls2\_2, nn\_ls2\_3))) & \rmse{0.43} \tabularnewline
    & 2. $\composec$(nn\_ls2\_4, $\composec$($\mapl$(nn\_ls2\_5), $\mapl$(nn\_ls2\_6))) & \rmse{0.45} \tabularnewline
    & 3. $\composec$($\composec$(nn\_ls2\_7, $\mapl$(nn\_ls2\_8)), $\mapl$(nn\_ls2\_9)) & \rmse{0.48} \tabularnewline
    \hline
    \multirow{3}{*}{Task 2: \countdigit{0}}
    & 1. $\composec$(nn\_ls2\_10, $\composec$($\mapl$(nn\_ls2\_11), $\mapl$(nn\_ls2\_12))) & \rmse{0.96} \tabularnewline
    & 2. $\composec$(lib.nn\_ls2\_1, $\mapl$($\composec$(nn\_ls2\_13, nn\_ls2\_14))) & \rmse{0.84} \tabularnewline
    & 3. $\composec$($\composec$(lib.nn\_ls2\_1, $\mapl$(nn\_ls2\_15)), $\mapl$(nn\_ls2\_16)) & \rmse{0.92} \tabularnewline
    \hline
    \multirow{3}{*}{Task 3: \istoy{1}}
    & 1. $\composec$(nn\_ls2\_17, nn\_ls2\_18) & \error{5.05} \tabularnewline
    & 2. $\composec$(nn\_ls2\_19, lib.nn\_ls2\_12) & \error{4.00} \tabularnewline
    & 3. $\composec$(lib.nn\_ls2\_2, nn\_ls2\_20) & \error{10.52} \tabularnewline
    \hline
    \multirow{3}{*}{Task 4: \recogdigit{5}}
    & 1. $\composec$(nn\_ls2\_21, nn\_ls2\_22) & \error{0.76} \tabularnewline
    & 2. $\composec$(nn\_ls2\_23, lib.nn\_ls2\_3) & \error{0.86} \tabularnewline
    & 3. $\composec$(lib.nn\_ls2\_17, nn\_ls2\_24) & \error{0.81} \tabularnewline
    \hline
    \multirow{3}{*}{Task 5: \countdigit{4}}
    & 1. $\composec$(lib.nn\_ls2\_1, $\mapl$($\composec$(nn\_ls2\_25, nn\_ls2\_26))) & \rmse{1.68} \tabularnewline
    & 2. $\composec$(lib.nn\_ls2\_1, $\mapl$($\composec$(nn\_ls2\_27, lib.nn\_ls2\_18))) & \rmse{1.51} \tabularnewline
    & 3. $\composec$(lib.nn\_ls2\_1, $\mapl$($\composec$(nn\_ls2\_28, lib.nn\_ls2\_12))) & \rmse{1.46} \tabularnewline
    \hline
    \multirow{3}{*}{Task 6: \countdigit{5}}
    & 1. $\composec$(nn\_ls2\_29, $\mapl$($\composec$(nn\_ls2\_30, nn\_ls2\_31))) & \rmse{0.43} \tabularnewline
    & 2. $\composec$(nn\_ls2\_32, $\mapl$($\composec$(lib.nn\_ls2\_21, lib.nn\_ls2\_22))) & \rmse{0.43} \tabularnewline
    & 3. $\composec$(lib.nn\_ls2\_1, $\mapl$($\composec$(nn\_ls2\_33, lib.nn\_ls2\_22))) & \rmse{0.45} \tabularnewline
    \hline
    \multirow{3}{*}{Task 7: \countdigit{1}}
    & 1. $\composec$(nn\_ls2\_34, $\mapl$($\composec$(lib.nn\_ls2\_25, nn\_ls2\_35))) & \rmse{0.64} \tabularnewline
    & 2. $\composec$(nn\_ls2\_36, $\mapl$($\composec$(nn\_ls2\_37, nn\_ls2\_38))) & \rmse{0.74} \tabularnewline
    & 3. $\composec$(nn\_ls2\_39, $\mapl$($\composec$(nn\_ls2\_40, lib.nn\_ls2\_26))) & \rmse{0.83} \tabularnewline
    \hline
    \multirow{3}{*}{Task 8: \recogdigit{1}}
    & 1. $\composec$(nn\_ls2\_41, lib.nn\_ls2\_3) & \error{0.29} \tabularnewline
    & 2. $\composec$(nn\_ls2\_42, lib.nn\_ls2\_12) & \error{0.19} \tabularnewline
    & 3. $\composec$(nn\_ls2\_43, lib.nn\_ls2\_22) & \error{0.24} \tabularnewline
    \hline
    \multirow{3}{*}{Task 9: \countdigit{8}}
    & 1. $\composec$(nn\_ls2\_44, $\mapl$($\composec$(nn\_ls2\_45, lib.nn\_ls2\_31))) & \rmse{0.46} \tabularnewline
    & 2. $\composec$(nn\_ls2\_46, $\mapl$($\composec$(nn\_ls2\_47, lib.nn\_ls2\_26))) & \rmse{0.45} \tabularnewline
    & 3. $\composec$(nn\_ls2\_48, $\mapl$($\composec$(nn\_ls2\_49, lib.nn\_ls2\_3))) & \rmse{0.47} \tabularnewline
\end{tabular}
}\end{table*}

\begin{table*}[]{\small \centering \scriptsize
\caption{Long Sequence 3(LS3).}
\label{tab:LS3Progs}
\begin{tabular}{l>{\raggedright}p{\progcolwidth}r}
    \textbf{Task} & \textbf{Top 3 Programs} & \score \tabularnewline
    \hline
    \multirow{3}{*}{Task 1: \countdigit{9}}
    & 1. $\composec$(nn\_ls3\_1, $\composec$($\mapl$(nn\_ls3\_2), $\mapl$(nn\_ls3\_3))) & \rmse{0.46} \tabularnewline
    & 2. $\composec$(nn\_ls3\_4, $\mapl$($\composec$(nn\_ls3\_5, nn\_ls3\_6))) & \rmse{0.48} \tabularnewline
    & 3. $\composec$($\composec$(nn\_ls3\_7, $\mapl$(nn\_ls3\_8)), $\mapl$(nn\_ls3\_9)) & \rmse{0.55} \tabularnewline
    \hline
    \multirow{3}{*}{Task 2: \countdigit{1}}
    & 1. $\composec$(lib.nn\_ls3\_1, $\mapl$($\composec$(nn\_ls3\_10, nn\_ls3\_11))) & \rmse{0.63} \tabularnewline
    & 2. $\composec$(lib.nn\_ls3\_1, $\composec$($\mapl$(nn\_ls3\_12), $\mapl$(nn\_ls3\_13))) & \rmse{0.68} \tabularnewline
    & 3. $\composec$($\composec$(lib.nn\_ls3\_1, $\mapl$(nn\_ls3\_14)), $\mapl$(nn\_ls3\_15)) & \rmse{0.63} \tabularnewline
    \hline
    \multirow{3}{*}{Task 3: \istoy{2}}
    & 1. $\composec$(nn\_ls3\_16, nn\_ls3\_17) & \error{8.19} \tabularnewline
    & 2. $\composec$(nn\_ls3\_18, lib.nn\_ls3\_11) & \error{9.95} \tabularnewline
    & 3. $\composec$(lib.nn\_ls3\_2, nn\_ls3\_19) & \error{14.00} \tabularnewline
    \hline
    \multirow{3}{*}{Task 4: \recogdigit{1}}
    & 1. $\composec$(nn\_ls3\_20, lib.nn\_ls3\_3) & \error{0.38} \tabularnewline
    & 2. $\composec$(nn\_ls3\_21, lib.nn\_ls3\_17) & \error{0.48} \tabularnewline
    & 3. $\composec$(nn\_ls3\_22, nn\_ls3\_23) & \error{0.24} \tabularnewline
    \hline
    \multirow{3}{*}{Task 5: \countdigit{3}}
    & 1. $\composec$(lib.nn\_ls3\_1, $\mapl$($\composec$(nn\_ls3\_24, nn\_ls3\_25))) & \rmse{0.51} \tabularnewline
    & 2. $\composec$(lib.nn\_ls3\_1, $\mapl$($\composec$(nn\_ls3\_26, lib.nn\_ls3\_17))) & \rmse{0.66} \tabularnewline
    & 3. $\composec$(lib.nn\_ls3\_1, $\mapl$($\composec$(nn\_ls3\_27, lib.nn\_ls3\_11))) & \rmse{0.61} \tabularnewline
    \hline
    \multirow{3}{*}{Task 6: \countdigit{1}}
    & 1. $\composec$(nn\_ls3\_28, $\mapl$($\composec$(nn\_ls3\_29, lib.nn\_ls3\_11))) & \rmse{0.38} \tabularnewline
    & 2. $\composec$(nn\_ls3\_30, $\mapl$($\composec$(nn\_ls3\_31, lib.nn\_ls3\_3))) & \rmse{0.37} \tabularnewline
    & 3. $\composec$(lib.nn\_ls3\_1, $\mapl$($\composec$(nn\_ls3\_32, lib.nn\_ls3\_3))) & \rmse{0.40} \tabularnewline
    \hline
    \multirow{3}{*}{Task 7: \countdigit{2}}
    & 1. $\composec$(nn\_ls3\_33, $\mapl$($\composec$(nn\_ls3\_34, lib.nn\_ls3\_17))) & \rmse{0.96} \tabularnewline
    & 2. $\composec$(lib.nn\_ls3\_1, $\mapl$($\composec$(nn\_ls3\_35, nn\_ls3\_36))) & \rmse{0.99} \tabularnewline
    & 3. $\composec$(lib.nn\_ls3\_1, $\mapl$($\composec$(nn\_ls3\_37, lib.nn\_ls3\_17))) & \rmse{0.90} \tabularnewline
    \hline
    \multirow{3}{*}{Task 8: \recogdigit{9}}
    & 1. $\composec$(nn\_ls3\_38, nn\_ls3\_39) & \error{1.52} \tabularnewline
    & 2. $\composec$(lib.nn\_ls3\_2, nn\_ls3\_40) & \error{2.43} \tabularnewline
    & 3. $\composec$(lib.nn\_ls3\_2, lib.nn\_ls3\_3) & \error{1.43} \tabularnewline
    \hline
    \multirow{3}{*}{Task 9: \countdigit{3}}
    & 1. $\composec$(nn\_ls3\_41, $\mapl$($\composec$(nn\_ls3\_42, nn\_ls3\_43))) & \rmse{0.39} \tabularnewline
    & 2. $\composec$(nn\_ls3\_44, $\mapl$($\composec$(nn\_ls3\_45, lib.nn\_ls3\_39))) & \rmse{0.42} \tabularnewline
    & 3. $\composec$(nn\_ls3\_46, $\mapl$($\composec$(nn\_ls3\_47, lib.nn\_ls3\_3))) & \rmse{0.44} \tabularnewline
\end{tabular}
}\end{table*}

\begin{table*}[]{\small \centering \scriptsize
\caption{Long Sequence 4(LS4).}
\label{tab:LS4Progs}
\begin{tabular}{l>{\raggedright}p{\progcolwidth}r}
    \textbf{Task} & \textbf{Top 3 Programs} & \score \tabularnewline
    \hline
    \multirow{3}{*}{Task 1: \countdigit{6}}
    & 1. $\composec$(nn\_ls4\_1, $\composec$($\mapl$(nn\_ls4\_2), $\mapl$(nn\_ls4\_3))) & \rmse{0.40} \tabularnewline
    & 2. $\composec$(nn\_ls4\_4, $\mapl$($\composec$(nn\_ls4\_5, nn\_ls4\_6))) & \rmse{0.45} \tabularnewline
    & 3. $\composec$($\composec$(nn\_ls4\_7, $\mapl$(nn\_ls4\_8)), $\mapl$(nn\_ls4\_9)) & \rmse{0.48} \tabularnewline
    \hline
    \multirow{3}{*}{Task 2: \countdigit{2}}
    & 1. $\composec$(lib.nn\_ls4\_1, $\composec$($\mapl$(nn\_ls4\_10), $\mapl$(nn\_ls4\_11))) & \rmse{0.89} \tabularnewline
    & 2. $\composec$(lib.nn\_ls4\_1, $\mapl$($\composec$(nn\_ls4\_12, nn\_ls4\_13))) & \rmse{0.99} \tabularnewline
    & 3. $\composec$($\composec$(lib.nn\_ls4\_1, $\mapl$(nn\_ls4\_14)), $\mapl$(nn\_ls4\_15)) & \rmse{0.91} \tabularnewline
    \hline
    \multirow{3}{*}{Task 3: \istoy{3}}
    & 1. $\composec$(nn\_ls4\_16, lib.nn\_ls4\_11) & \error{4.95} \tabularnewline
    & 2. $\composec$(nn\_ls4\_17, nn\_ls4\_18) & \error{4.00} \tabularnewline
    & 3. $\composec$(lib.nn\_ls4\_10, nn\_ls4\_19) & \error{2.43} \tabularnewline
    \hline
    \multirow{3}{*}{Task 4: \recogdigit{8}}
    & 1. $\composec$(nn\_ls4\_20, lib.nn\_ls4\_3) & \error{0.71} \tabularnewline
    & 2. $\composec$(nn\_ls4\_21, nn\_ls4\_22) & \error{0.52} \tabularnewline
    & 3. $\composec$(nn\_ls4\_23, lib.nn\_ls4\_11) & \error{0.86} \tabularnewline
    \hline
    \multirow{3}{*}{Task 5: \countdigit{1}}
    & 1. $\composec$(lib.nn\_ls4\_1, $\mapl$($\composec$(nn\_ls4\_24, nn\_ls4\_25))) & \rmse{0.64} \tabularnewline
    & 2. $\composec$(lib.nn\_ls4\_1, $\mapl$($\composec$(nn\_ls4\_26, lib.nn\_ls4\_11))) & \rmse{0.57} \tabularnewline
    & 3. $\composec$(lib.nn\_ls4\_1, $\mapl$($\composec$(lib.nn\_ls4\_16, nn\_ls4\_27))) & \rmse{0.70} \tabularnewline
    \hline
    \multirow{3}{*}{Task 6: \countdigit{8}}
    & 1. $\composec$(lib.nn\_ls4\_1, $\mapl$($\composec$(nn\_ls4\_28, lib.nn\_ls4\_3))) & \rmse{0.39} \tabularnewline
    & 2. $\composec$(lib.nn\_ls4\_1, $\mapl$($\composec$(nn\_ls4\_29, nn\_ls4\_30))) & \rmse{0.38} \tabularnewline
    & 3. $\composec$(lib.nn\_ls4\_1, $\mapl$($\composec$(nn\_ls4\_31, lib.nn\_ls4\_11))) & \rmse{0.40} \tabularnewline
    \hline
    \multirow{3}{*}{Task 7: \countdigit{3}}
    & 1. $\composec$(lib.nn\_ls4\_1, $\mapl$($\composec$(nn\_ls4\_32, lib.nn\_ls4\_11))) & \rmse{0.61} \tabularnewline
    & 2. $\composec$(lib.nn\_ls4\_1, $\mapl$($\composec$(nn\_ls4\_33, nn\_ls4\_34))) & \rmse{0.54} \tabularnewline
    & 3. $\composec$(lib.nn\_ls4\_1, $\mapl$($\composec$(nn\_ls4\_35, lib.nn\_ls4\_25))) & \rmse{0.60} \tabularnewline
    \hline
    \multirow{3}{*}{Task 8: \recogdigit{6}}
    & 1. $\composec$(nn\_ls4\_36, lib.nn\_ls4\_3) & \error{0.81} \tabularnewline
    & 2. $\composec$(lib.nn\_ls4\_20, nn\_ls4\_37) & \error{0.90} \tabularnewline
    & 3. $\composec$(nn\_ls4\_38, nn\_ls4\_39) & \error{0.86} \tabularnewline
    \hline
    \multirow{3}{*}{Task 9: \countdigit{5}}
    & 1. $\composec$(lib.nn\_ls4\_1, $\mapl$($\composec$(nn\_ls4\_40, nn\_ls4\_41))) & \rmse{0.37} \tabularnewline
    & 2. $\composec$(lib.nn\_ls4\_1, $\mapl$($\composec$(nn\_ls4\_42, lib.nn\_ls4\_3))) & \rmse{0.39} \tabularnewline
    & 3. $\composec$(lib.nn\_ls4\_1, $\mapl$($\composec$(nn\_ls4\_43, lib.nn\_ls4\_11))) & \rmse{0.39} \tabularnewline
\end{tabular}
}\end{table*}

\begin{table*}[]{\small \centering \scriptsize
\caption{Long Sequence 5(LS5).}
\label{tab:LS5Progs}
\begin{tabular}{l>{\raggedright}p{\progcolwidth}r}
    \textbf{Task} & \textbf{Top 3 Programs} & \score \tabularnewline
    \hline
    \multirow{3}{*}{Task 1: \countdigit{4}}
    & 1. $\composec$(nn\_ls5\_1, $\composec$($\mapl$(nn\_ls5\_2), $\mapl$(nn\_ls5\_3))) & \rmse{0.45} \tabularnewline
    & 2. $\composec$(nn\_ls5\_4, $\mapl$($\composec$(nn\_ls5\_5, nn\_ls5\_6))) & \rmse{0.46} \tabularnewline
    & 3. $\composec$($\composec$(nn\_ls5\_7, $\mapl$(nn\_ls5\_8)), $\mapl$(nn\_ls5\_9)) & \rmse{0.48} \tabularnewline
    \hline
    \multirow{3}{*}{Task 2: \countdigit{3}}
    & 1. $\composec$(nn\_ls5\_10, $\composec$($\mapl$(nn\_ls5\_11), $\mapl$(lib.nn\_ls5\_3))) & \rmse{0.60} \tabularnewline
    & 2. $\composec$(lib.nn\_ls5\_1, $\composec$($\mapl$(nn\_ls5\_12), $\mapl$(nn\_ls5\_13))) & \rmse{0.63} \tabularnewline
    & 3. $\composec$($\composec$(lib.nn\_ls5\_1, $\mapl$(nn\_ls5\_14)), $\mapl$(nn\_ls5\_15)) & \rmse{0.58} \tabularnewline
    \hline
    \multirow{3}{*}{Task 3: \istoy{4}}
    & 1. $\composec$(nn\_ls5\_16, nn\_ls5\_17) & \error{20.33} \tabularnewline
    & 2. $\composec$(lib.nn\_ls5\_11, nn\_ls5\_18) & \error{17.76} \tabularnewline
    & 3. $\composec$(nn\_ls5\_19, lib.nn\_ls5\_3) & \error{21.38} \tabularnewline
    \hline
    \multirow{3}{*}{Task 4: \recogdigit{7}}
    & 1. $\composec$(nn\_ls5\_20, nn\_ls5\_21) & \error{1.19} \tabularnewline
    & 2. $\composec$(nn\_ls5\_22, lib.nn\_ls5\_3) & \error{0.90} \tabularnewline
    & 3. $\composec$(lib.nn\_ls5\_2, nn\_ls5\_23) & \error{1.62} \tabularnewline
    \hline
    \multirow{3}{*}{Task 5: \countdigit{0}}
    & 1. $\composec$(lib.nn\_ls5\_10, $\mapl$($\composec$(nn\_ls5\_24, nn\_ls5\_25))) & \rmse{0.90} \tabularnewline
    & 2. $\composec$(lib.nn\_ls5\_10, $\mapl$($\composec$(nn\_ls5\_26, lib.nn\_ls5\_17))) & \rmse{0.90} \tabularnewline
    & 3. $\composec$(lib.nn\_ls5\_10, $\mapl$($\composec$(nn\_ls5\_27, lib.nn\_ls5\_3))) & \rmse{0.86} \tabularnewline
    \hline
    \multirow{3}{*}{Task 6: \countdigit{7}}
    & 1. $\composec$(lib.nn\_ls5\_1, $\mapl$($\composec$(nn\_ls5\_28, nn\_ls5\_29))) & \rmse{0.47} \tabularnewline
    & 2. $\composec$(lib.nn\_ls5\_1, $\mapl$($\composec$(nn\_ls5\_30, lib.nn\_ls5\_21))) & \rmse{0.47} \tabularnewline
    & 3. $\composec$(nn\_ls5\_31, $\mapl$($\composec$(lib.nn\_ls5\_16, nn\_ls5\_32))) & \rmse{0.47} \tabularnewline
    \hline
    \multirow{3}{*}{Task 7: \countdigit{4}}
    & 1. $\composec$(nn\_ls5\_33, $\mapl$($\composec$(nn\_ls5\_34, lib.nn\_ls5\_25))) & \rmse{1.72} \tabularnewline
    & 2. $\composec$(nn\_ls5\_35, $\mapl$($\composec$(nn\_ls5\_36, lib.nn\_ls5\_17))) & \rmse{1.50} \tabularnewline
    & 3. $\composec$(lib.nn\_ls5\_1, $\mapl$($\composec$(nn\_ls5\_37, nn\_ls5\_38))) & \rmse{1.80} \tabularnewline
    \hline
    \multirow{3}{*}{Task 8: \recogdigit{4}}
    & 1. $\composec$(nn\_ls5\_39, lib.nn\_ls5\_3) & \error{0.29} \tabularnewline
    & 2. $\composec$(nn\_ls5\_40, lib.nn\_ls5\_21) & \error{0.38} \tabularnewline
    & 3. $\composec$(lib.nn\_ls5\_20, nn\_ls5\_41) & \error{0.48} \tabularnewline
    \hline
    \multirow{3}{*}{Task 9: \countdigit{0}}
    & 1. $\composec$(nn\_ls5\_42, $\mapl$($\composec$(nn\_ls5\_43, nn\_ls5\_44))) & \rmse{0.37} \tabularnewline
    & 2. $\composec$(nn\_ls5\_45, $\mapl$($\composec$(lib.nn\_ls5\_24, nn\_ls5\_46))) & \rmse{0.40} \tabularnewline
    & 3. $\composec$(nn\_ls5\_47, $\mapl$($\composec$(nn\_ls5\_48, lib.nn\_ls5\_21))) & \rmse{0.40} \tabularnewline
\end{tabular}
}\end{table*}

\begin{table*}[]
{\scriptsize \centering 
\caption{Counting Sequence 1(CS1), Evolutionary Algorithm. ``CE'' denotes classification error and ``RMSE'' denotes
root mean square error. }
\label{tab:CS1ProgsEvo}
\begin{tabular}{lp{\progcolwidth}r}
    \textbf{Task} & \textbf{Top 3 programs} & \score \tabularnewline
    \hline
    \multirow{3}{*}{Task 1: \recogdigit{d_1}}
    & 1. $\composec$(nn\_cs1\_1, nn\_cs1\_2) & \error{0.57} \tabularnewline
    & 2. $\composec$(nn\_cs1\_3, nn\_cs1\_2) & \error{0.38} \tabularnewline
    & 3. $\composec$(nn\_cs1\_4, nn\_cs1\_2) & \error{0.76} \tabularnewline
    \hline
    \multirow{3}{*}{Task 2: \recogdigit{d_2}}
    & 1. $\composec$(nn\_cs1\_5, nn\_cs1\_6) & \error{0.38} \tabularnewline
    & 2. $\composec$(nn\_cs1\_7, nn\_cs1\_8) & \error{0.48} \tabularnewline
    & 3. $\composec$(nn\_cs1\_9, nn\_cs1\_10) & \error{0.43} \tabularnewline
    \hline
    \multirow{3}{*}{Task 3: \countdigit{d_1}}
    & 1. $\composec$(nn\_cs1\_11, $\mapl$($\composec$(nn\_cs1\_12, lib.nn\_cs1\_2))) & \rmse{0.38} \tabularnewline
    & 2. $\composec$(nn\_cs1\_13, $\mapl$($\composec$(nn\_cs1\_14, lib.nn\_cs1\_2))) & \rmse{0.38} \tabularnewline
    & 3. $\composec$(nn\_cs1\_15, $\mapl$($\composec$(lib.nn\_cs1\_1, lib.nn\_cs1\_2))) & \rmse{0.40} \tabularnewline
    \hline
    {Task 4: \countdigit{d_2}} & No Solution & \tabularnewline
\end{tabular}
}
\end{table*}

\begin{table*}[]{\small\centering \scriptsize
\caption{Counting Sequence 2(CS2), Evolutionary Algorithm.}
\label{tab:CS2ProgsEvo}
\begin{tabular}{lp{\progcolwidth}r}
    \textbf{Task} & \textbf{Top 3 programs} & \score \tabularnewline
    \hline
    \multirow{3}{*}{Task 1: \recogdigit{d_1}}
    & 1. $\composec$(nn\_cs2\_1, nn\_cs2\_2) & \error{1} \tabularnewline
    & 2. $\composec$(nn\_cs2\_3, nn\_cs2\_4) & \error{1} \tabularnewline
    & 3. $\composec$(nn\_cs2\_5, nn\_cs2\_2) & \error{1} \tabularnewline
    \hline
    \multirow{3}{*}{Task 2: \countdigit{d_1}}
    & 1. $\composec$(nn\_cs2\_6, $\mapl$($\composec$(lib.nn\_cs2\_1, lib.nn\_cs2\_2))) & \rmse{0.38} \tabularnewline
    & 2. $\composec$(nn\_cs2\_6, $\mapl$($\composec$(nn\_cs2\_7, lib.nn\_cs2\_2))) & \rmse{0.38} \tabularnewline
    & 3. $\composec$(nn\_cs2\_8, $\mapl$($\composec$(nn\_cs2\_9, nn\_cs2\_10))) & \rmse{0.39} \tabularnewline
    \hline
    Task 3: \countdigit{d_2}
    & No Solution & \tabularnewline
    \hline
    \multirow{3}{*}{Task 4: \recogdigit{d_2}}
    & 1. $\composec$(nn\_cs2\_11, nn\_cs2\_12) & \error{1} \tabularnewline
    & 2. $\composec$(nn\_cs2\_13, nn\_cs2\_14) & \error{1} \tabularnewline
    & 3. $\composec$(lib.nn\_cs2\_1, nn\_cs2\_15) & \error{1} \tabularnewline
\end{tabular}
}\end{table*}

\begin{table*}[]{\small \centering \scriptsize
\caption{Counting Sequence 3(CS3), Evolutionary Algorithm.}
\label{tab:CS3ProgsEvo}
\begin{tabular}{lp{\progcolwidth}r}
    \textbf{Task} & \textbf{Top 3 Programs} & \score \tabularnewline
    \hline
    \multirow{2}{*}{Task 1: \recogdigit{d}}
    & 1. $\composec$(nn\_cs3\_1, nn\_cs3\_2) & \error{0.57} \tabularnewline
    & 2. $\composec$(nn\_cs3\_3, nn\_cs3\_4) & \error{0.67} \tabularnewline
    & 3. $\composec$(nn\_cs3\_5, nn\_cs3\_6) & \error{0.62} \tabularnewline
    \hline
    \multirow{2}{*}{Task 2: \countdigit{d}}
    & 1. $\composec$(nn\_cs3\_7, $\mapl$($\composec$(nn\_cs3\_8, lib.nn\_cs3\_2))) & \rmse{0.36} \tabularnewline
    & 2. $\composec$(nn\_cs3\_7, $\mapl$($\composec$(nn\_cs3\_9, nn\_cs3\_10))) & \rmse{0.39} \tabularnewline
    & 3. $\composec$(nn\_cs3\_11, $\mapl$($\composec$(nn\_cs3\_12, nn\_cs3\_13))) & \rmse{0.39} \tabularnewline
    \hline
    \multirow{3}{*}{Task 3: \counttoy{t}}
    & 1. $\composec$(lib.nn\_cs3\_7, $\mapl$($\composec$(nn\_cs3\_14, nn\_cs3\_15))) & \rmse{0.70} \tabularnewline
    & 2. $\composec$(lib.nn\_cs3\_7, $\mapl$($\composec$(nn\_cs3\_16, nn\_cs3\_17))) & \rmse{0.61} \tabularnewline
    & 3. $\composec$(lib.nn\_cs3\_7, $\mapl$($\composec$(nn\_cs3\_18, nn\_cs3\_19))) & \rmse{0.64} \tabularnewline
    \hline
    \multirow{3}{*}{Task 4: \istoy{t}}
    & 1. $\composec$(nn\_cs3\_20, lib.nn\_cs3\_15) & \error{5.62} \tabularnewline
    & 2. $\composec$(lib.nn\_cs3\_14, lib.nn\_cs3\_15) & \error{5.38} \tabularnewline
    & 3. $\composec$(nn\_cs3\_21, lib.nn\_cs3\_15) & \error{5.76} \tabularnewline
\end{tabular}
}\end{table*}

\begin{table*}[]
{\small
\centering
\scriptsize
\caption{Summing Sequence(SS), Evolutionary Algorithm}
\label{tab:SSProgsEvo}
\begin{tabular}{l>{\raggedright}p{\progcolwidth}r}
    \textbf{Task} & \textbf{Top 3 programs} & \score \tabularnewline
    \hline
    Task 1: \classifydigit & 1. $\composec$(nn\_ss\_1, nn\_ss\_2) & \error{1}\tabularnewline
    \hline
    \multirow{2}{*}{Task 2: \sumdigits} & 1. $\composec$(($\foldl$ nn\_ss\_3 zeros(1)), $\mapl$($\composec$(nn\_ss\_4, lib.nn\_ss\_2)))
    & \rmse{6.64} \tabularnewline
    & 2. $\composec$(($\foldl$ nn\_ss\_5 zeros(1)), $\mapl$($\composec$(nn\_ss\_6, lib.nn\_ss\_2))) & \rmse{6.66} \tabularnewline
    & 3. $\composec$(($\foldl$ nn\_ss\_7 zeros(1)), $\mapl$($\composec$(nn\_ss\_8, lib.nn\_ss\_2))) & \rmse{6.70} \tabularnewline
\end{tabular}
}
\end{table*}

\begin{table*}[]{\small \centering \scriptsize
  \caption{Graph Sequence 1(GS1), Evolutionary Algorithm.}
  \label{tab:GS1ProgsEvo}
  \begin{tabular}{l>{\raggedright}p{\progcolwidth}r}
    \textbf{Task} & \textbf{Top 3 Programs} & \score \tabularnewline
\hline
\multirow{1}{*}{Task 1: \regressspeed}
& 1. $\composec$(nn\_gs1\_1, nn\_gs1\_2) & \rmse{0.80} \tabularnewline
\hline
\multirow{3}{*}{Task 2: \shortestpathstreet}
& 1. $\mapg$($\composec$(nn\_gs1\_3, lib.nn\_gs1\_2)) & \rmse{8.36} \tabularnewline
& 2. $\mapg$($\composec$(nn\_gs1\_4, nn\_gs1\_5)) & \rmse{8.37} \tabularnewline
& 3. $\mapg$($\composec$(nn\_gs1\_6, lib.nn\_gs1\_2)) & \rmse{8.35} \tabularnewline
   \end{tabular}
}\end{table*}


\begin{table*}[]{\small \centering \scriptsize
  \caption{Graph Sequence 2(GS2), Evolutionary Algorithm.}
  \label{tab:GS2ProgsEvo}
  \begin{tabular}{l>{\raggedright}p{\progcolwidth}r}
    \textbf{Task} & \textbf{Top 3 Programs} & \score \tabularnewline
\hline
\multirow{1}{*}{Task 1: \regressmnist}
& 1. $\composec$(nn\_gs2\_1, nn\_gs2\_2) & \rmse{1.47} \tabularnewline
\hline
\multirow{3}{*}{Task 2: \shortestpathmnist}
& 1. $\mapg$($\composec$(lib.nn\_gs2\_1, nn\_gs2\_3)) & \rmse{6.58} \tabularnewline
& 2. $\mapg$($\composec$(lib.nn\_gs2\_1, nn\_gs2\_4)) & \rmse{6.59} \tabularnewline
& 3. $\mapg$($\composec$(lib.nn\_gs2\_1, nn\_gs2\_5)) & \rmse{6.63} \tabularnewline
\hline
\multirow{3}{*}{Task 3: \shortestpathstreet}
& 1. $\mapg$($\composec$(lib.nn\_gs2\_1, nn\_gs2\_6)) & \rmse{7.82} \tabularnewline
& 2. $\mapg$($\composec$(lib.nn\_gs2\_1, nn\_gs2\_7)) & \rmse{7.87} \tabularnewline
& 3. $\mapg$($\composec$(nn\_gs2\_8, nn\_gs2\_9)) & \rmse{7.96} \tabularnewline
   \end{tabular}
}\end{table*}


\begin{table*}[]{\small \centering \scriptsize
\caption{Long Sequence 1(LS1), Evolutionary Algorithm.}
\label{tab:LS1ProgsEvo}
\begin{tabular}{lp{\progcolwidth}r}
    \textbf{Task} & \textbf{Top 3 Programs} & \score \tabularnewline
    \hline
    \multirow{3}{*}{Task 1: \countdigit{7}}
    & 1. $\composec$(nn\_ls1\_1, $\mapl$($\composec$(nn\_ls1\_2, nn\_ls1\_3))) & \rmse{0.42} \tabularnewline
    & 2. $\composec$(nn\_ls1\_4, $\mapl$($\composec$(nn\_ls1\_5, nn\_ls1\_6))) & \rmse{0.44} \tabularnewline
    & 3. $\composec$(nn\_ls1\_1, $\mapl$($\composec$(nn\_ls1\_7, nn\_ls1\_8))) & \rmse{0.50} \tabularnewline
    \hline
    \multirow{3}{*}{Task 2: \countdigit{4}}
    & 1. $\composec$(lib.nn\_ls1\_1, $\mapl$($\composec$(nn\_ls1\_9, nn\_ls1\_10))) & \rmse{1.65} \tabularnewline
    & 2. $\composec$(lib.nn\_ls1\_1, $\mapl$($\composec$(nn\_ls1\_11, nn\_ls1\_12))) & \rmse{1.53} \tabularnewline
    & 3. $\composec$(lib.nn\_ls1\_1, $\mapl$($\composec$(nn\_ls1\_13, nn\_ls1\_14))) & \rmse{1.60} \tabularnewline
    \hline
    \multirow{3}{*}{Task 3: \istoy{0}}
    & 1. $\composec$(nn\_ls1\_15, lib.nn\_ls1\_10) & \error{9.81} \tabularnewline
    & 2. $\composec$(nn\_ls1\_16, nn\_ls1\_17) & \error{9.76} \tabularnewline
    & 3. $\composec$(nn\_ls1\_18, lib.nn\_ls1\_10) & \error{8.76} \tabularnewline
    \hline
    \multirow{3}{*}{Task 4: \recogdigit{9}}
    & 1. $\composec$(nn\_ls1\_19, nn\_ls1\_20) & \error{1.43} \tabularnewline
    & 2. $\composec$(nn\_ls1\_21, nn\_ls1\_22) & \error{1.43} \tabularnewline
    & 3. $\composec$(nn\_ls1\_23, nn\_ls1\_24) & \error{1.62} \tabularnewline
    \hline
    \multirow{3}{*}{Task 5: \countdigit{2}}
    & 1. $\composec$(nn\_ls1\_25, $\mapl$($\composec$(lib.nn\_ls1\_19, nn\_ls1\_26))) & \rmse{0.90} \tabularnewline
    & 2. $\composec$(lib.nn\_ls1\_1, $\mapl$($\composec$(nn\_ls1\_27, nn\_ls1\_28))) & \rmse{0.94} \tabularnewline
    & 3. $\composec$(lib.nn\_ls1\_1, $\mapl$($\composec$(nn\_ls1\_29, nn\_ls1\_30))) & \rmse{0.98} \tabularnewline
    \hline
    Task 6: \countdigit{9}
    & No Solution & \tabularnewline
    \hline
    Task 7: \countdigit{0}
    & No Solution & \tabularnewline
    \hline
    \multirow{3}{*}{Task 8: \recogdigit{7}}
    & 1. $\composec$(nn\_ls1\_31, lib.nn\_ls1\_26) & \error{1.29} \tabularnewline
    & 2. $\composec$(nn\_ls1\_32, lib.nn\_ls1\_3) & \error{0.71} \tabularnewline
    & 3. $\composec$(nn\_ls1\_33, nn\_ls1\_34) & \error{1.19} \tabularnewline
    \hline
    \multirow{3}{*}{Task 9: \countdigit{2}}
    & 1. $\composec$(nn\_ls1\_35, $\mapl$($\composec$(nn\_ls1\_36, lib.nn\_ls1\_3))) & \rmse{0.36} \tabularnewline
    & 2. $\composec$(lib.nn\_ls1\_25, $\mapl$($\composec$(nn\_ls1\_36, nn\_ls1\_37))) & \rmse{0.38} \tabularnewline
    & 3. $\composec$(nn\_ls1\_38, $\mapl$($\composec$(nn\_ls1\_39, nn\_ls1\_40))) & \rmse{0.37} \tabularnewline
\end{tabular}
}\end{table*}

\begin{table*}[]{\small \centering \scriptsize
\caption{Long Sequence 2(LS2), Evolutionary Algorithm.}
\label{tab:LS2ProgsEvo}
\begin{tabular}{lp{\progcolwidth}r}
    \textbf{Task} & \textbf{Top 3 Programs} & \score \tabularnewline
    \hline
    Task 1: \countdigit{1}
    & No Solution & \tabularnewline
    \hline
    Task 2: \countdigit{0}
    & No Solution & \tabularnewline
    \hline
    \multirow{3}{*}{Task 3: \istoy{1}}
    & 1. $\composec$(nn\_ls2\_1, nn\_ls2\_2) & \error{5.43} \tabularnewline
    & 2. $\composec$(nn\_ls2\_3, nn\_ls2\_4) & \error{5.81} \tabularnewline
    & 3. $\composec$(nn\_ls2\_5, nn\_ls2\_6) & \error{5.05} \tabularnewline
    \hline
    \multirow{3}{*}{Task 4: \recogdigit{5}}
    & 1. $\composec$(nn\_ls2\_7, nn\_ls2\_8) & \error{0.71} \tabularnewline
    & 2. $\composec$(nn\_ls2\_9, nn\_ls2\_10) & \error{0.43} \tabularnewline
    & 3. $\composec$(nn\_ls2\_9, nn\_ls2\_11) & \error{0.62} \tabularnewline
    \hline
    Task 5: \countdigit{4}
    & No Solution & \tabularnewline
    \hline
    Task 6: \countdigit{5}
    & No Solution & \tabularnewline
    \hline
    Task 7: \countdigit{1}
    & No Solution & \tabularnewline
    \hline
    \multirow{3}{*}{Task 8: \recogdigit{1}}
    & 1. $\composec$(nn\_ls2\_12, nn\_ls2\_13) & \error{0.19} \tabularnewline
    & 2. $\composec$(nn\_ls2\_14, lib.nn\_ls2\_2) & \error{0.29} \tabularnewline
    & 3. $\composec$(nn\_ls2\_15, lib.nn\_ls2\_2) & \error{0.33} \tabularnewline
    \hline
    Task 9: \countdigit{8}
    & No Solution & \tabularnewline
\end{tabular}
}\end{table*}

\begin{table*}[]{\small \centering \scriptsize
\caption{Long Sequence 3(LS3), Evolutionary Algorithm.}
\label{tab:LS3ProgsEvo}
\begin{tabular}{lp{\progcolwidth}r}
    \textbf{Task} & \textbf{Top 3 Programs} & \score \tabularnewline
    \hline
    Task 1: \countdigit{9}
    & No Solution & \tabularnewline
    \hline
    Task 2: \countdigit{1}
    & No Solution & \tabularnewline
    \hline
    \multirow{3}{*}{Task 3: \istoy{2}}
    & 1. $\composec$(nn\_ls3\_1, nn\_ls3\_2) & \error{10.52} \tabularnewline
    & 2. $\composec$(nn\_ls3\_3, nn\_ls3\_2) & \error{9.14} \tabularnewline
    & 3. $\composec$(nn\_ls3\_4, nn\_ls3\_2) & \error{10.81} \tabularnewline
    \hline
    \multirow{3}{*}{Task 4: \recogdigit{1}}
    & 1. $\composec$(nn\_ls3\_5, nn\_ls3\_6) & \error{0.48} \tabularnewline
    & 2. $\composec$(nn\_ls3\_7, lib.nn\_ls3\_2) & \error{0.33} \tabularnewline
    & 3. $\composec$(nn\_ls3\_8, nn\_ls3\_9) & \error{0.24} \tabularnewline
    \hline
    Task 5: \countdigit{3}
    & No Solution & \tabularnewline
    \hline
    \multirow{3}{*}{Task 6: \countdigit{1}}
    & 1. $\composec$(nn\_ls3\_10, $\mapl$($\composec$(nn\_ls3\_11, lib.nn\_ls3\_6))) & \rmse{0.38} \tabularnewline
    & 2. $\composec$(nn\_ls3\_12, $\mapl$($\composec$(nn\_ls3\_13, nn\_ls3\_14))) & \rmse{0.37} \tabularnewline
    & 3. $\composec$(nn\_ls3\_15, $\mapl$($\composec$(nn\_ls3\_11, lib.nn\_ls3\_6))) & \rmse{0.39} \tabularnewline
    \hline
    \multirow{3}{*}{Task 7: \countdigit{2}}
    & 1. $\composec$(lib.nn\_ls3\_10, $\mapl$($\composec$(nn\_ls3\_16, nn\_ls3\_17))) & \rmse{1.02} \tabularnewline
    & 2. $\composec$(lib.nn\_ls3\_10, $\mapl$($\composec$(nn\_ls3\_18, nn\_ls3\_17))) & \rmse{0.92} \tabularnewline
    & 3. $\composec$(lib.nn\_ls3\_10, $\mapl$($\composec$(nn\_ls3\_16, nn\_ls3\_19))) & \rmse{0.96} \tabularnewline
    \hline
    \multirow{3}{*}{Task 8: \recogdigit{9}}
    & 1. $\composec$(nn\_ls3\_20, nn\_ls3\_21) & \error{1.05} \tabularnewline
    & 2. $\composec$(nn\_ls3\_22, nn\_ls3\_23) & \error{1.14} \tabularnewline
    & 3. $\composec$(nn\_ls3\_24, lib.nn\_ls3\_17) & \error{1.86} \tabularnewline
    \hline
    \multirow{3}{*}{Task 9: \countdigit{3}}
    & 1. $\composec$(lib.nn\_ls3\_10, $\mapl$($\composec$(nn\_ls3\_25, lib.nn\_ls3\_21))) & \rmse{0.45} \tabularnewline
    & 2. $\composec$(lib.nn\_ls3\_10, $\mapl$($\composec$(nn\_ls3\_26, lib.nn\_ls3\_17))) & \rmse{0.47} \tabularnewline
    & 3. $\composec$(lib.nn\_ls3\_10, $\mapl$($\composec$(nn\_ls3\_25, lib.nn\_ls3\_21))) & \rmse{0.46} \tabularnewline
\end{tabular}
}\end{table*}

\begin{table*}[]{\small \centering \scriptsize
\caption{Long Sequence 4(LS4), Evolutionary Algorithm.}
\label{tab:LS4ProgsEvo}
\begin{tabular}{lp{\progcolwidth}r}
    \textbf{Task} & \textbf{Top 3 Programs} & \score \tabularnewline
    \hline
    Task 1: \countdigit{6}
    & No Solution & \tabularnewline
    \hline
    Task 2: \countdigit{2}
    & No Solution & \tabularnewline
    \hline
    \multirow{3}{*}{Task 3: \istoy{3}}
    & 1. $\composec$(nn\_ls4\_1, nn\_ls4\_2) & \error{5.10} \tabularnewline
    & 2. $\composec$(nn\_ls4\_3, nn\_ls4\_4) & \error{3.57} \tabularnewline
    & 3. $\composec$(nn\_ls4\_5, nn\_ls4\_6) & \error{4.24} \tabularnewline
    \hline
    \multirow{3}{*}{Task 4: \recogdigit{8}}
    & 1. $\composec$(nn\_ls4\_7, nn\_ls4\_8) & \error{0.33} \tabularnewline
    & 2. $\composec$(nn\_ls4\_9, nn\_ls4\_10) & \error{0.48} \tabularnewline
    & 3. $\composec$(nn\_ls4\_11, nn\_ls4\_12) & \error{0.90} \tabularnewline
    \hline
    Task 5: \countdigit{1}
    & No Solution & \tabularnewline
    \hline
    \multirow{3}{*}{Task 6: \countdigit{8}}
    & 1. $\composec$(nn\_ls4\_13, $\mapl$($\composec$(nn\_ls4\_14, nn\_ls4\_15))) & \rmse{0.41} \tabularnewline
    & 2. $\composec$(nn\_ls4\_16, $\mapl$($\composec$(nn\_ls4\_17, lib.nn\_ls4\_8))) & \rmse{0.44} \tabularnewline
    & 3. $\composec$(nn\_ls4\_18, $\mapl$($\composec$(nn\_ls4\_19, lib.nn\_ls4\_8))) & \rmse{0.44} \tabularnewline
    \hline
    \multirow{3}{*}{Task 7: \countdigit{3}}
    & 1. $\composec$(lib.nn\_ls4\_13, $\mapl$($\composec$(nn\_ls4\_20, nn\_ls4\_21))) & \rmse{0.56} \tabularnewline
    & 2. $\composec$(lib.nn\_ls4\_13, $\mapl$($\composec$(nn\_ls4\_22, lib.nn\_ls4\_2))) & \rmse{0.59} \tabularnewline
    & 3. $\composec$(lib.nn\_ls4\_13, $\mapl$($\composec$(nn\_ls4\_20, lib.nn\_ls4\_2))) & \rmse{0.57} \tabularnewline
    \hline
    \multirow{3}{*}{Task 8: \recogdigit{6}}
    & 1. $\composec$(nn\_ls4\_23, lib.nn\_ls4\_15) & \error{0.48} \tabularnewline
    & 2. $\composec$(nn\_ls4\_24, lib.nn\_ls4\_15) & \error{0.62} \tabularnewline
    & 3. $\composec$(nn\_ls4\_25, lib.nn\_ls4\_15) & \error{0.71} \tabularnewline
    \hline
    \multirow{3}{*}{Task 9: \countdigit{5}}
    & 1. $\composec$(lib.nn\_ls4\_13, $\mapl$($\composec$(nn\_ls4\_26, nn\_ls4\_27))) & \rmse{0.41} \tabularnewline
    & 2. $\composec$(lib.nn\_ls4\_13, $\mapl$($\composec$(nn\_ls4\_26, nn\_ls4\_27))) & \rmse{0.41} \tabularnewline
    & 3. $\composec$(lib.nn\_ls4\_13, $\mapl$($\composec$(nn\_ls4\_28, nn\_ls4\_29))) & \rmse{0.41} \tabularnewline
\end{tabular}
}\end{table*}

\begin{table*}[]{\small \centering \scriptsize
\caption{Long Sequence 5(LS5), Evolutionary Algorithm.}
\label{tab:LS5ProgsEvo}
\begin{tabular}{lp{\progcolwidth}r}
    \textbf{Task} & \textbf{Top 3 Programs} & \score \tabularnewline
    \hline
    Task 1: \countdigit{4}
    & No Solution & \tabularnewline
    \hline
    Task 2: \countdigit{3}
    & No Solution & \tabularnewline
    \hline
    \multirow{3}{*}{Task 3: \istoy{4}}
    & 1. $\composec$(nn\_ls5\_1, nn\_ls5\_2) & \error{17.00} \tabularnewline
    & 2. $\composec$(nn\_ls5\_3, nn\_ls5\_4) & \error{21.62} \tabularnewline
    & 3. $\composec$(nn\_ls5\_5, nn\_ls5\_6) & \error{16.52} \tabularnewline
    \hline
    \multirow{3}{*}{Task 4: \recogdigit{7}}
    & 1. $\composec$(nn\_ls5\_7, nn\_ls5\_8) & \error{1.14} \tabularnewline
    & 2. $\composec$(nn\_ls5\_9, nn\_ls5\_10) & \error{0.95} \tabularnewline
    & 3. $\composec$(nn\_ls5\_11, nn\_ls5\_12) & \error{1.00} \tabularnewline
    \hline
    Task 5: \countdigit{0}
    & No Solution & \tabularnewline
    \hline
    Task 6: \countdigit{7}
    & No Solution & \tabularnewline
    \hline
    Task 7: \countdigit{4}
    & No Solution & \tabularnewline
    \hline
    \multirow{3}{*}{Task 8: \recogdigit{4}}
    & 1. $\composec$(nn\_ls5\_13, nn\_ls5\_14) & \error{0.38} \tabularnewline
    & 2. $\composec$(nn\_ls5\_15, lib.nn\_ls5\_8) & \error{0.33} \tabularnewline
    & 3. $\composec$(nn\_ls5\_15, nn\_ls5\_16) & \error{0.33} \tabularnewline
    \hline
    \multirow{3}{*}{Task 9: \countdigit{0}}
    & 1. $\composec$(nn\_ls5\_17, $\mapl$($\composec$(nn\_ls5\_18, lib.nn\_ls5\_8))) & \rmse{0.38} \tabularnewline
    & 2. $\composec$(nn\_ls5\_17, $\mapl$($\composec$(nn\_ls5\_19, lib.nn\_ls5\_2))) & \rmse{0.38} \tabularnewline
    & 3. $\composec$(nn\_ls5\_17, $\mapl$($\composec$(nn\_ls5\_20, lib.nn\_ls5\_8))) & \rmse{0.40} \tabularnewline
\end{tabular}
}\end{table*}

\section{Full Experimental Results on Counting Tasks}

\begin{figure*}[t!]
    \centering
    \begin{subfigure}[t]{2.4in}
        \centering
        \includegraphics[width=2.3in]{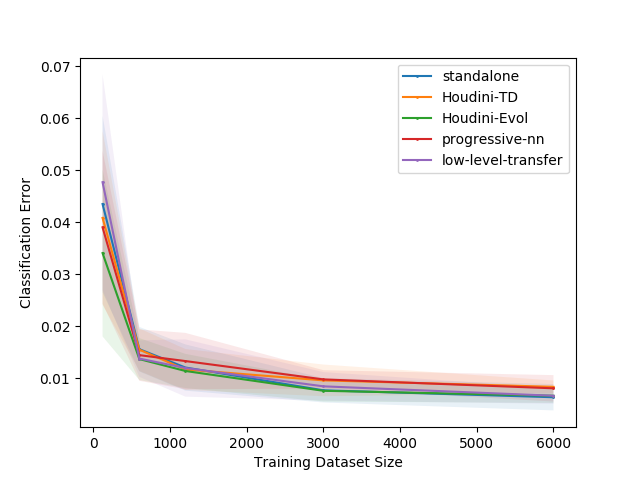}
        \caption{Task 1: \isdigit{d_1}}\label{fig:cs1t1appx}
    \end{subfigure}
    \begin{subfigure}[t]{2.4in}
        \centering
        \includegraphics[width=2.3in]{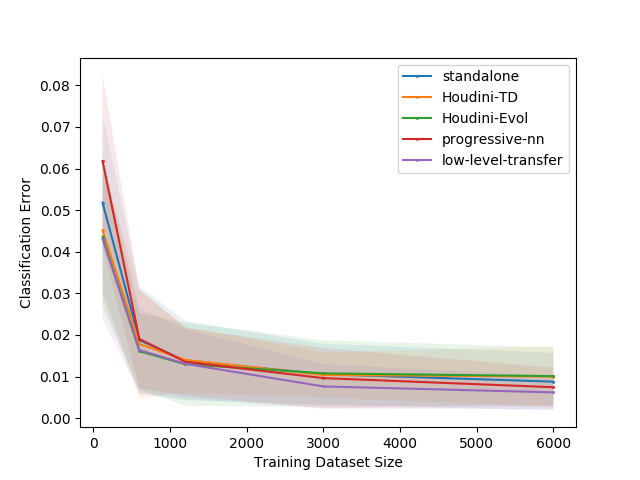}
        \caption{Task 2: \isdigit{d_2}}\label{fig:cs1t2appx}
    \end{subfigure}
    \begin{subfigure}[t]{2.4in}
        \centering
        \includegraphics[width=2.3in]{s1t3}
        \caption{Task 3: \countdigit{d_1}}\label{fig:cs1t3appx}
    \end{subfigure}
    \begin{subfigure}[t]{2.4in}
        \centering
        \includegraphics[width=2.3in]{s1t4}
        \caption{Task 4: \countdigit{d_2}}\label{fig:cs1t4appx}
    \end{subfigure}
    \caption{Lifelong learning for ``learning to count'' (Sequence CS1), demonstrating low-level transfer
    of perceptual recognizers.}\label{fig:cs1appx}
\end{figure*}

\begin{figure*}[t!]
    \centering
    \begin{subfigure}[t]{2.4in}
        \centering
        \includegraphics[width=2.3in]{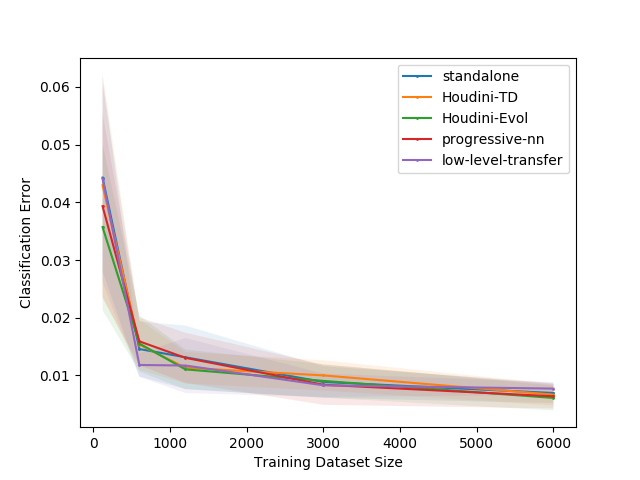}
        \caption{Task 1: \isdigit{d_1}}\label{fig:cs2t1appx}
    \end{subfigure}
    \begin{subfigure}[t]{2.4in}
        \centering
        \includegraphics[width=2.3in]{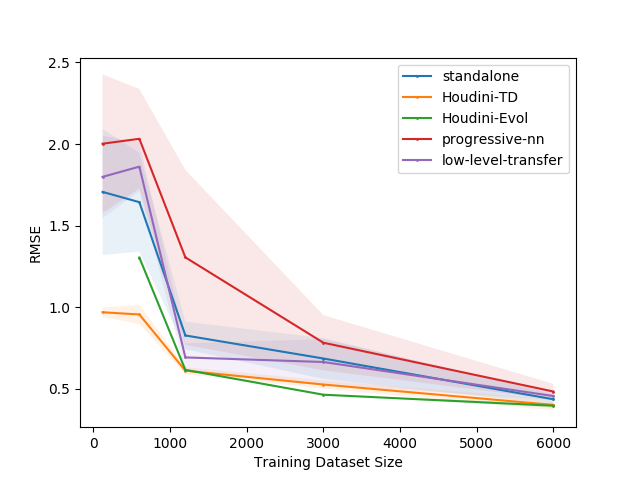}
        \caption{Task 2: \countdigit{d_1}}\label{fig:cs2t2appx}
    \end{subfigure}
    \begin{subfigure}[t]{2.4in}
        \centering
        \includegraphics[width=2.3in]{s2t3}
        \caption{Task 3: \countdigit{d_2}}\label{fig:cs2t3appx}
    \end{subfigure}
    \begin{subfigure}[t]{2.4in}
        \centering
        \includegraphics[width=2.3in]{s2t4}
        \caption{Task 4: \isdigit{d_2}}\label{fig:cs2t4appx}
    \end{subfigure}
    \caption{Lifelong learning for ``learning to count'' (Sequence CS2), demonstrating high-level transfer
    of a counting network across categories.}\label{fig:cs2appx}
\end{figure*}

\begin{figure*}[t!]
    \centering
    \begin{subfigure}[t]{2.4in}
        \centering
        \includegraphics[width=2.3in]{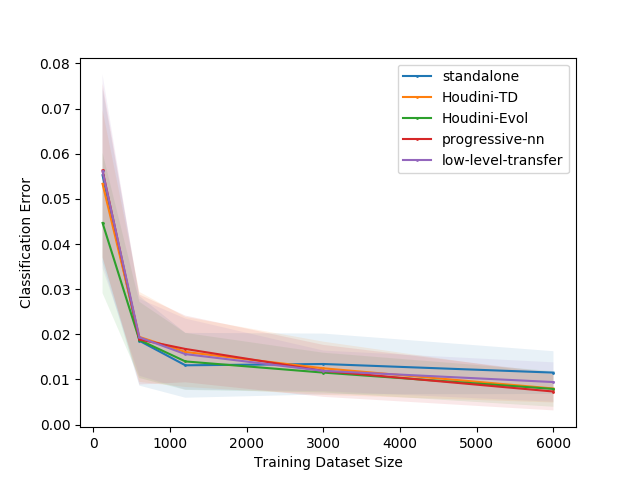}
        \caption{Task 1: \isdigit{d_1}}\label{fig:3t1appx}
    \end{subfigure}
    \begin{subfigure}[t]{2.4in}
        \centering
        \includegraphics[width=2.3in]{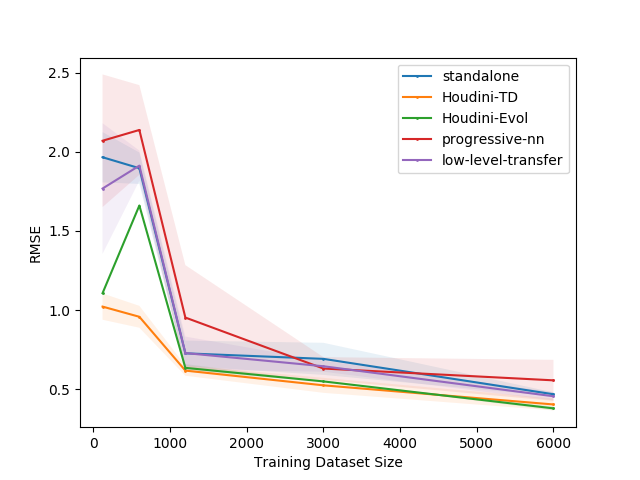}
        \caption{Task 2: \countdigit{d_1}}\label{fig:3t2appx}
    \end{subfigure}
    \begin{subfigure}[t]{2.4in}
        \centering
        \includegraphics[width=2.3in]{s3t3}
        \caption{Task 3: \counttoy{t_1}}\label{fig:3t3appx}
    \end{subfigure}
    \begin{subfigure}[t]{2.4in}
        \centering
        \includegraphics[width=2.3in]{s3t4}
        \caption{Task 4: \istoy{t_1}}\label{fig:3t4appx}
    \end{subfigure}
    \caption{Lifelong learning for ``learning to count'' (Sequence CS3), demonstrating high-level transfer across different
    types of images. After learning to count MNIST digits,
    the same network can be used to count images
    of toys.}\label{fig:cs3appx}
\end{figure*}

In the paper, we present results
for the counting sequences on for the
later tasks, in which transfer learning
is possible. For completeness, in this section
we present results on all of the tasks in the sequences. See Figures \ref{fig:cs1appx}--\ref{fig:cs3appx}.
We note that for the early tasks in each task sequence (e.g. CS1 tasks 1 and 2),
there is little relevant information that can be transferred from early tasks,
so as expected all methods perform similarly; e.g., the output of \system is a single library function.


\section{Results on Longer Task Sequence LS}

\begin{figure}[h!]
\includegraphics[width=0.45\textwidth]{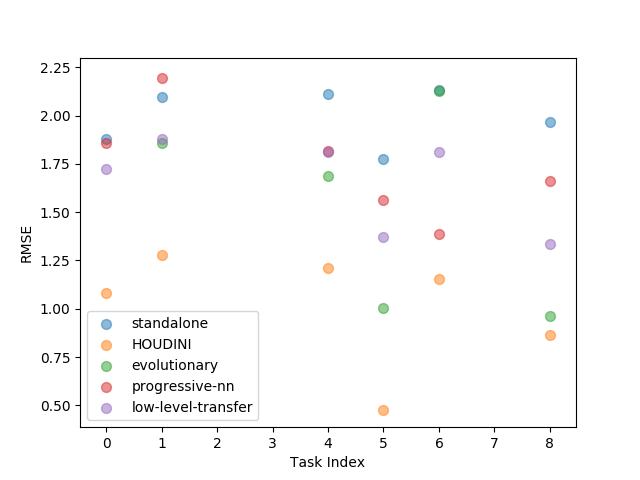}
\hspace{1em}
\includegraphics[width=0.45\textwidth]{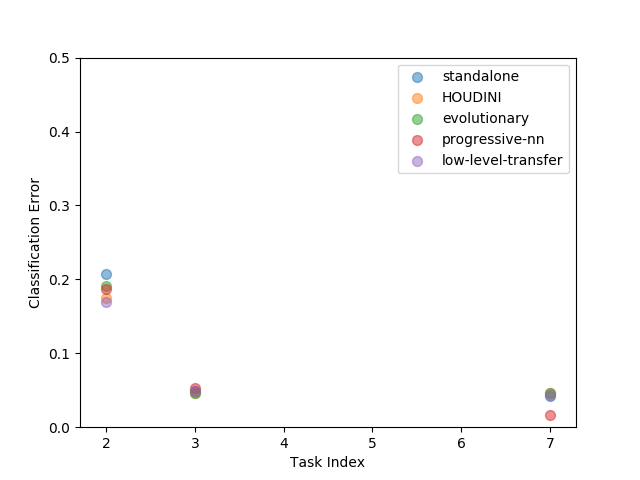}
\caption{Performance of transfer learning systems on task sequence LS1. On the left:
regression tasks. On the right: classification tasks}\label{fig:ls}
\end{figure}

We report the performance of all methods on the longer task sequences
on Figure~\ref{fig:ls}.
To save space, we report performance of all methods when trained
on 10\% of the data. The full learning curves follow similar
patterns as the other task sequences. We report the classification
and regression tasks from LS separately, because the error functions
for the two tasks have different dynamic ranges. 
Please note that in the Figure, the tasks are labelled starting from $0$.
On the classification tasks, we note that all methods have similar
performance. Examining the task sequence LS from Figure \ref{fig:tasks}, we see that these
tasks have no opportunity to transfer from earlier tasks.
On the regression tasks however, there is opportunity
to transfer, and we see that \system shows much better performance
than the other methods.

\end{document}